# Monte Carlo Sampling Methods for Approximating Interactive POMDPs

**Prashant Doshi**        PDOSHI@CS.UGA.EDU
*Department of Computer Science*
*University of Georgia*
*415 Boyd GSRC*
*Athens, GA 30602*

**Piotr J. Gmytrasiewicz**        PIOTR@CS.UIC.EDU
*Department of Computer Science*
*University of Illinois at Chicago*
*851 S. Morgan St*
*Chicago, IL 60607*

## Abstract

Partially observable Markov decision processes (POMDPs) provide a principled framework for sequential planning in uncertain single agent settings. An extension of POMDPs to multiagent settings, called interactive POMDPs (I-POMDPs), replaces POMDP belief spaces with interactive hierarchical belief systems which represent an agent's belief about the physical world, about beliefs of other agents, and about their beliefs about others' beliefs. This modification makes the difficulties of obtaining solutions due to complexity of the belief and policy spaces even more acute. We describe a general method for obtaining approximate solutions of I-POMDPs based on particle filtering (PF). We introduce the *interactive PF*, which descends the levels of the interactive belief hierarchies and samples and propagates beliefs at each level. The interactive PF is able to mitigate the belief space complexity, but it does not address the policy space complexity. To mitigate the policy space complexity – sometimes also called the curse of history – we utilize a complementary method based on sampling likely observations while building the look ahead reachability tree. While this approach does not completely address the curse of history, it beats back the curse's impact substantially. We provide experimental results and chart future work.

## 1. Introduction

Interactive POMDPs (I-POMDPs) (Gmytrasiewicz & Doshi, 2005; Seuken & Zilberstein, 2008) are a generalization of POMDPs to multiagent settings and offer a principled decision-theoretic framework for sequential decision making in uncertain multiagent settings. I-POMDPs are applicable to autonomous self-interested agents who locally compute what actions they should execute to optimize their preferences given what they believe while interacting with others with possibly conflicting objectives. Though POMDPs can be used in multiagent settings, it is so only under the strong assumption that the other agent's behavior be adequately represented implicitly (say, as noise) within the POMDP model (see Boutilier, Dean, & Hanks, 1999; Gmytrasiewicz & Doshi, 2005, for examples). The approach adopted in I-POMDPs is to expand the traditional state space to include models of other agents. Some of these models are the sophisticated *intentional* models, which ascribe beliefs, preferences, and rationality to others and are analogous to the notion of agent types in Bayesian games (Harsanyi, 1967; Mertens & Zamir, 1985). Other models, such as finite state machines, do not ascribe beliefs or rationality to other agents and we call them *subintentional* models. An agent's beliefs within I-POMDPs are called interactive beliefs, and they are nested analogously to the hierarchical belief systems considered in game theory (Mertens & Zamir, 1985; Brandenburger & Dekel, 1993; Heifetz & Samet, 1998; Aumann, 1999), in theoretical computer science (Fagin, Halpern, Moses, & Vardi, 1995) and to the hyper-priors in hierarchical Bayesian models (Gelman, Carlin, Stern, & Rubin, 2004). Since the interactive beliefs may be infinitely nested, Gmytrasiewicz and Doshi (2005) defined finitely nested I-POMDPs as computable specializations of the infinitely nested ones. Solutions of finitely nested I-POMDPs map an agent's states of belief about the environment and other agents' models to policies. Consequently, I-POMDPs find important applications in agent, human, and mixed agent-human environments. Some potential applications include path planning in multi-robot environments, coordinating troop movements in battlefields, planning the course of a treatment in a multi-treatment therapy, and explaining commonly observed social behaviors (Doshi, Zeng, & Chen, 2007).

However, optimal decision making in uncertain multiagent settings is computationally very hard requiring significant time and memory resources. For example, the problem of solving decentralized POMDPs has been shown to lie in the NEXP-complete class (Bernstein, Givan, Immerman, & Zilberstein, 2002). Expectedly, exact solutions of finitely nested I-POMDPs are difficult to compute as well, due to two primary sources of intractability: ($i$) The complexity of the belief representation which is proportional to the dimensions of the belief simplex, sometimes called the *curse of dimensionality*. ($ii$) The complexity of the space of policies, which is proportional to the number of possible future beliefs, also called the *curse of history*.

Both these sources of intractability exist in POMDPs also (see Pineau, Gordon, & Thrun, 2006; Poupart & Boutilier, 2004) but the curse of dimensionality is especially more acute in I-POMDPs. This is because in I-POMDPs the complexity of the belief space is even greater; the beliefs may include beliefs about the physical environment, and possibly the agent's beliefs about other agents' beliefs, about their beliefs about others', and so on. Thus, a contributing factor to the curse of dimensionality is the level of belief nesting that is considered. As the total number of agent models grows exponentially with the increase in nesting level, so does the solution complexity.

We observe that one approach to solving a finitely nested I-POMDP is to investigate collapsing the model to a traditional POMDP, and utilize available approximation methods that apply to POMDPs. However, the transformation into a POMDP is not straightforward. In particular, it does not seem possible to model the update of other agents' nested beliefs as a part of the transition function in the POMDP. Such a transition function would include nested beliefs and require solutions of others' models in defining it, and thus be quite different from the standard ones to which current POMDP approaches apply.

In this article, we present the first set of generally applicable methods for computing *approximately* optimal policies for the finitely nested I-POMDP framework while demonstrating computational savings. Since an agent's belief is defined over other agents' models, which may be a complex infinite space, sampling methods which are able to approximate distributions over large spaces to arbitrary accuracy are a promising approach. We adopt the particle filter (Gordon, Salmond, & Smith, 1993; Doucet, Freitas, & Gordon, 2001) as our point of departure. There is growing empirical evidence (Koller & Lerner, 2001; Daum & Huang, 2002) that particle filters are unable to significantly reduce the adverse impact of increasing state spaces. Specifically, the number of particles needed to maintain the error from the exact state estimation increases as the number of dimensions increase.

However, the rate of convergence of the approximate posterior to the true one is independent of the dimensions of the state space (Crisan & Doucet, 2002) under weak assumptions. In other words, while we may need more particles to maintain error as the state space increases, the rate at which the error reduces remains unchanged, regardless of the state space. Furthermore, sampling approaches allow us to *focus* resources on the regions of the state space that are considered more likely in an uncertain environment, providing a strong potential for computational savings.

We generalize the particle filter, and more specifically the bootstrap filter (Gordon et al., 1993), to the multiagent setting, resulting in the *interactive particle filter* (I-PF). The generalization is not trivial: We do not simply treat the other agent as an automaton whose actions follow a fixed and known distribution. Rather, we consider the case where other agents are intentional – they possess beliefs, capabilities and preferences. Subsequently, the propagation step in the I-PF becomes more complicated than in the standard PF. In projecting the subject agent's belief over time, we must project the other agent's belief, which involves predicting its action and anticipating its observations. Mirroring the hierarchical character of interactive beliefs, the interactive particle filtering involves sampling and propagation at each of the hierarchical levels of the beliefs. We empirically demonstrate the ability of the I-PF to flexibly approximate the state estimation in I-POMDPs, and show the computational savings obtained in comparison to a regular grid based implementation. However, as we sample an identical number of particles at each nesting level, the total number of particles and the associated complexity, continues to grow exponentially with the nesting level.

We combine the I-PF with value iteration on sample sets thereby providing a general way to solve finitely nested I-POMDPs. Our approximation method is *anytime* and is applicable to agents that start with a prior belief and optimize over finite horizons. Consequently, our method finds applications for online plan computation. We derive error bounds for our approach that are applicable to singly-nested I-POMDPs and discuss the difficulty in generalizing the bounds to multiply nested beliefs. We empirically demonstrate the performance and computational savings obtained by our method on standard test problems as well as a larger uninhabited aerial vehicle (UAV) reconnaissance problem.

While the I-PF is able to flexibly mitigate the belief space complexity, it does not address the policy space complexity. In order to mitigate the curse of history, we present a complementary method based on sampling observations while building the look ahead reachability tree during value iteration. This translates into considering only those future beliefs during value iteration that an agent is likely to have from a given belief. This approach is similar in spirit to the sparse sampling techniques used in generating partial look ahead trees for action selection during reinforcement learning (Kearns, Mansour, & Ng, 2002; Wang, Lizotte, Bowling, & Schuurmans, 2005) and for online planning in POMDPs (Ross, Pineau, Paquet, & Chaib-draa, 2008). While these approaches were applied in single agent reinforcement learning problems, we focus on a multiagent setting and recursively apply the technique to solve models of all agents at each nesting level. Observation sampling was also recently utilized in DEC-POMDPs (Seuken & Zilberstein, 2007), where it was shown to improve the performance on large problems. We note that this approach does not completely address the curse of history, but beats back its impact on the difficulty of computing the I-POMDP solutions, substantially. We report on the additional computational savings obtained when we combine this method with the I-PF, and provide empirical results in support.

Rest of this article is structured in the following manner. We review the various state estimation methods and their relevance, and the use of particle filters in previous works in Section 2. In Section 3, we review the traditional particle filtering technique concentrating on bootstrap filters in particular. We briefly outline the finitely nested I-POMDP framework in Section 4 and the multi-agent tiger problem used for illustration in Section 5. In Section 6, we discuss representations for the nested beliefs and the inherent difficulty in formulating them. In order to facilitate understanding, we give a decomposition of the I-POMDP belief update in Section 7. We then present the I-PF that approximates the finitely nested I-POMDP belief update in Section 8. This is followed by a method that utilizes the I-PF to compute solutions to I-POMDPs, in Section 9. We also comment on the asymptotic convergence and compute error bounds of our approach. In Section 10, we report on the performance of our approximation method on simple and larger test problems. In Section 11, we provide a technique for mitigating the curse of history, and report on some empirical results. Finally, we conclude this article and outline future research directions in Section 12.

## 2. Related Work

Several approaches to nonlinear Bayesian estimation exist. Among these, the extended Kalman filter (EKF) (Sorenson, 1985), is most popular. The EKF linearises the estimation problem so that the Kalman filter can be applied. The required probability density function (p.d.f.) is still approximated by a Gaussian, which may lead to filter divergence, and therefore an increase in the error. Other approaches include the Gaussian sum filter (Sorenson & Alspach, 1971), and superimposing a grid over the state space with the belief being evaluated only over the grid points (Kramer & Sorenson, 1988). In the latter approach, the choice of an efficient grid is non-trivial, and the method suffers from the curse of dimensionality: The number of grid points that must be considered is exponential in the dimensions of the state space. Recently, techniques that utilize Monte Carlo (MC) sampling for approximating the Bayesian state estimation problem have received much attention. These techniques are general enough, in that, they are applicable to both linear, as well as, non-linear problem dynamics, and the rate of convergence of the approximation error to zero is independent of the dimensions of the underlying state space. Among the spectrum of MC techniques, two that have been particularly well-studied in sequential settings are Markov chain Monte Carlo (MCMC) (Hastings, 1970; Gelman et al., 2004), and particle filters (Gordon et al., 1993; Doucet et al., 2001). Approximating the I-POMDP belief update using the former technique, may turn out to be computationally exhaustive. Specifically, MCMC algorithms that utilize rejection sampling (e.g. Hastings, 1970) may cause a large number of intentional models to be sampled, solved, and rejected, before one is utilized for propagation. In addition, the complex estimation process in I-POMDPs makes the task of computing the acceptance ratio for rejection sampling computationally inefficient. Although Gibbs sampling (Gelman et al., 2004) avoids rejecting samples, it would involve sampling from a conditional distribution of the physical state given the observation history and model of other, and from the distribution of the other's model given the physical state. However, these distributions are neither efficient to compute nor easy to derive analytically. Particle filters need not reject solved models and compute a new model in replacement, propagating all solved models over time and resampling them. They are intuitively amenable to approximating the I-POMDP belief update and produce reasonable approximations of the posterior while being computationally feasible.

Particle filters previously have been successfully applied to approximate the belief update in continuous state space single agent POMDPs (Thrun, 2000; Poupart, Ortiz, & Boutilier, 2001). While Thrun (2000) integrates particle filtering with Q-learning to *learn* the policy, Poupart et al. (2001) assume the prior existence of an exact value function and present an error bound analysis of substituting the POMDP belief update with particle filters. Loosely related to our work are the sampling algorithms that appear in (Ortiz & Kaelbling, 2000) for selecting actions in influence diagrams, but this work does not focus on sequential decision making. In the multiagent setting, particle filters have been employed for collaborative multi-robot localization (Fox, Burgard, Kruppa, & Thrun, 2000). In this application, the emphasis was on predicting the position of the robot, and not the actions of the other robots, which is a critical step in our approach. Additionally, to facilitate fast localization, beliefs of other robots encountered during motion were considered to be fully observable to enable synchronization.

Within the POMDP literature, approaches other than sampling methods have also appeared that address the curse of dimensionality. An important class of such algorithms prescribe substituting the complex belief space with a simpler subspace (Bertsekas, 1995; Tsitsiklis & Roy, 1996; Poupart & Boutilier, 2003; Roy, Gordon, & Thrun, 2005). The premise of these methods is that the beliefs – distributions over all the physical states – contain more information than required in order to plan near-optimally. Poupart and Boutilier (2003) use Krylov subspaces (Saad, 1996) to directly compress the POMDP model, and analyze the effect of the compression on the decision quality. To ensure lossless compression, i.e. the decision quality at each compressed belief is not compromised, the transition and reward functions must be linear. Roy et al. (2005) proposed using principal component analysis (Collins, Dasgupta, & R.E.Schapire, 2002) to uncover a low dimensional belief subspace that usually encompasses a robot's potential beliefs. The method is based on the observation that beliefs along many real-world trajectories exhibit only a few degrees of freedom. The effectiveness of these methods is problem specific; indeed, it is possible to encounter problems where no substantial belief compression may occur. When applied to the I-POMDP framework, the effectiveness of the compression techniques would depend, for example, on the existence of agent models whose likelihoods within the agent's belief do not change after successive belief updates or on the existence of correlated agent models. Whether such models exist in practice is a topic of future work.

Techniques that address the curse of history in POMDPs also exist. Poupart and Boutilier (2004) generate policies via policy iteration using finite state controllers with a bounded number of nodes. Pineau et al. (2006) perform point-based value iteration (PBVI) by selecting a small subset of reachable belief points at each step from the belief simplex and planning only over these belief points. Doshi and Perez (2008) outline the challenges and develop PBVI for I-POMDPs. Though our method of mitigating the curse of history is conceptually close to point based selection methods, we focus on plan computation when the initial belief is known while the previously mentioned methods are typically utilized for offline planning. An approximate way of solving POMDPs online is the RTBSS approach (Paquet, Tobin, & Chaib-draa, 2005; Ross et al., 2008) that adopts the branch-and-bound technique for pruning the look ahead reachability tree. This approach focuses on selecting the best action to expand which is complementary to our approach of sampling the observations. Further, its extension to the multiagent setting as formalized by I-POMDPs may not be trivial due to the need for a bounding heuristic function whose formulation in multiagent settings remains to be investigated.

## 3. Background: Particle Filter for the Single Agent Setting

To act rationally in uncertain settings, agents need to track the evolution of the state over time, based on the actions they perform and the available observations. In single agent settings, the state estimation is usually accomplished with a technique called the *Bayes filter* (Russell & Norvig, 2003). A Bayes filter allows the agent to maintain a belief about the state of the world at any given time, and update this belief each time an action is performed and new sensory information arrives. The convenience of this approach lies in the fact that the update is independent of the past percepts and action sequences. This is because the agent's belief is a *sufficient statistic*: it fully summarizes all of the information contained in past actions and observations.

The operation of a Bayes filter can be decomposed into a two-step process:

- *Prediction:* When an agent performs a new action, $a^{t-1}$, its prior belief state is updated:

$$Pr(s^t|a^{t-1}, b^{t-1}) = \int_{s^{t-1}} b^{t-1}(s^{t-1}) T(s^t|s^{t-1}, a^{t-1}) ds^{t-1} \qquad (1)$$

- *Correction:* Thereafter, when an observation, $o^t$, is received, the intermediate belief state, $Pr(\cdot|a^{t-1}, b^{t-1})$, is corrected:

$$Pr(s^t|o^t, a^{t-1}, b^{t-1}) = \alpha O(o^t|s^t, a^{t-1}) Pr(s^t|a^{t-1}, b^{t-1}) \qquad (2)$$

where $\alpha$ is the normalizing constant, $T$ is the transition function that gives the uncertain effect of performing an action on the physical state, and $O$ is the observation function which gives the likelihood of receiving an observation from a state on performing an action.

Particle filters (PF) (Gordon et al., 1993; Doucet et al., 2001) are specific implementations of Bayes filters tailored toward making Bayes filters applicable to non-linear dynamic systems. Rather than sampling directly from the target distribution which is often difficult, PFs adopt the method of importance sampling (Geweke, 1989), which allows samples to be drawn from a more tractable distribution called the *proposal distribution*, $\pi$. For example, if $Pr(S^t|o^t, a^{t-1}, b^{t-1})$ is the target posterior distribution, and $\pi(S^t|o^t, a^{t-1}, b^{t-1})$ the proposal distribution, and the support of $\pi(S^t|o^t, a^{t-1}, b^{t-1})$ includes the support of $Pr(S^t|o^t, a^{t-1}, b^{t-1})$, we can approximate the target posterior by sampling $N$ i.i.d. particles $\{s^{(n)}, n = 1...N\}$ according to $\pi(S^t|o^t, a^{t-1}, b^{t-1})$ and assigning to each particle a normalized importance weight:

$$w^{(n)} = \frac{\widetilde{w}(s^{(n)})}{\sum_{n=1}^{N} \widetilde{w}(s^{(n)})} \quad \text{where} \quad \widetilde{w}(s^{(n)}) = \frac{Pr(s^{(n)}|o^t, a^{t-1}, b^{t-1})}{\pi(s^{(n)}|o^t, a^{t-1}, b^{t-1})}$$

Each true probability, $Pr(s|o^t, a^{t-1}, b^{t-1})$, is then approximated by:

$$Pr_N(s|o^t, a^{t-1}, b^{t-1}) = \sum_{n=1}^{N} w^{(n)} \delta_D(s - s^{(n)})$$

where $\delta_D(\cdot)$ is the Dirac-delta function. As $N \to \infty$, $Pr_N(s|o^t, a^{t-1}, b^{t-1}) \overset{a.s.}{\to} Pr(s|o^t, a^{t-1}, b^{t-1})$. When applied recursively over several steps, importance sampling leads to a large variance in the weights. To avoid this degeneracy, Gordon et al. (1993) suggested inserting a resampling step, which would increase the population of those particles that had high importance weights. This has the beneficial effect of focusing the particles in the high likelihood regions supported by the observations and increasing the tracking ability of the PF. Since particle filtering extends importance sampling sequentially and appends a resampling step, it has also been called sequential importance sampling and resampling (SISR).

The general algorithm for the particle filtering technique is given by Doucet et al. (2001). We concentrate on a specific implementation of this algorithm, that has previously been studied under various names such as MC localization, survival of the fittest, and the bootstrap filter. The implementation maintains a set of $N$ particles denoted by $\tilde{b}^{t-1}$ independently sampled from the prior, $b^{t-1}$, and takes an action and observation as input. Each particle is then propagated forwards in time, using the transition kernel $T$ of the environment. Each particle is then weighted by the likelihood of perceiving the observation from the state that the particle represents, as given by the observation function $O$. This is followed by the (unbiased) resampling step, in which particles are picked proportionately to their weights, and a uniform weight is subsequently attached to each particle. We outline the algorithm of the bootstrap filter in Fig. 1. Crisan and Doucet (2002) outline a rigorous proof of the convergence of this algorithm toward the true posterior as $N \rightarrow \infty$.

---

**Function** PARTICLEFILTER($\tilde{b}^{t-1}, a^{t-1}, o^t$) **returns** $\tilde{b}^t$

1. $\tilde{b}^{tmp} \leftarrow \phi, \tilde{b}^t \leftarrow \phi$

   *Importance Sampling*

2. **for all** $s^{(n),t-1} \in \tilde{b}^{t-1}$ **do**

3.     Sample $s^{(n),t} \sim T(S^t|a^{t-1}, s^{(n),t-1})$

4.     Weight $s^{(n),t}$ with the importance weight:
   $$\widetilde{w}^{(n)} = O(o^t|s^{(n),t}, a^{t-1})$$

5.     $\tilde{b}^{tmp} \xleftarrow{\cup} (s^{(n),t}, \widetilde{w}^{(n)})$

6. Normalize all $\widetilde{w}^{(n)}$ so that $\sum_{n=1}^N w^{(n)} = 1$

   _Selection_

7. Resample with replacement $N$ particles $\{s^{(n),t}, n = 1...N\}$ from the set $\tilde{b}^{tmp}$ according to the importance weights.

8. $\tilde{b}^t \leftarrow \{s^{(n),t}, n = 1...N\}$

9. **return** $\tilde{b}^t$

**end function**

---

Figure 1: The particle filtering algorithm for approximating the Bayes filter.

Let us understand the working of the PF in the context of a simple example – the single agent tiger problem (Kaelbling, Littman, & Cassandra, 1998). The single agent tiger problem resembles a game show in which the agent has to choose to open one of two doors behind which lies either a valuable prize or a dangerous tiger. Apart from actions that open doors, the subject has the option of listening for the tiger's growl coming from the left, or the right door. However, the subject's hearing is imperfect, with given percentages (say, 15%) of false positive and false negative occurrences. Following Kaelbling et al. (1998), we assume that the value of the prize is 10, that the pain associated with encountering the tiger can be quantified as -100, and that the cost of listening is -1.

Let the agent have a prior belief according to which it is uninformed about the location of the tiger. In other words, it believes with a probability of 0.5 that the tiger is behind the left door (TL), and with a similar probability that the tiger is behind the right door (TR). We will see how the agent approximately updates its belief using the particle filter when, say, it listens (L) and hears a growl from the left (GL). Fig. 2 illustrates the particle filtering process. Since the agent is uninformed about the tiger's location, we start with an equal number of particles (samples) denoting TL (lightly shaded) and TR (darkly shaded). The initial sample set is approximately representative of the agent's prior belief of 0.5. Since listening does not change the location of the tiger, the composition of the sample set remains unchanged after propagation. On hearing a growl from the left, the light particles denoting TL will be tagged with a larger weight (0.85) because they are more likely to be responsible for GL, than the dark particles denoting TR (0.15). Here, the size of the particle is proportional to the weight attached to the particle. Finally, the resampling step yields the sample set at time step $t$, which contains more particles denoting TL than TR. This sample set approximately represents the updated belief of 0.85 of the agent that the tiger is behind the left door. Note that the propagation carries out the task of prediction as shown in Eq. 1 approximately, while the correction step (Eq. 2) is approximately performed by weighting and resampling.

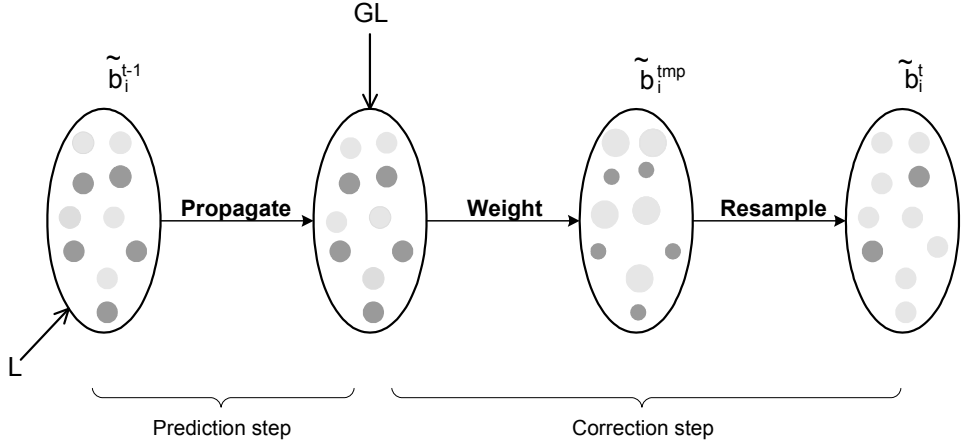

Figure 2: Particle filtering for state estimation in the single agent tiger problem. The light and dark particles denote the states TL and TR respectively. The particle filtering process consists of three steps: *Propagation* (line 3 of Fig. 1), *Weighting* (line 4), and *Resampling* (line 7).

## 4. Overview of Finitely Nested I-POMDPs

I-POMDPs (Gmytrasiewicz & Doshi, 2005) generalize POMDPs to handle multiple agents. They do this by including models of other agents in the state space. We focus on finitely nested I-POMDPs here, which are the computable counterparts of I-POMDPs in general. For simplicity of presentation let us consider an agent, $i$, that is interacting with one other agent, $j$. The arguments generalize to a setting with more than two agents in a straightforward manner.

**Definition 1** (I-POMDP$_{i,l}$). *A finitely nested interactive POMDP of agent $i$, I-POMDP$_{i,l}$, is:*

$$I\text{-}POMDP_{i,l} = \langle IS_{i,l}, A, T_i, \Omega_i, O_i, R_i \rangle$$

*where:*

- $IS_{i,l}$ is a set of **interactive** states defined as $IS_{i,l} = S \times M_{j,l-1}$, $l \geq 1$, and $IS_{i,0} = S$,[1] where $S$ is the set of states of the physical environment, and $M_{j,l-1}$ is the set of possible models of agent $j$.

---

1. If there are more agents participating in the interaction, $K > 2$, then $IS_{i,l} = S \times_{j=1}^{K-1} M_{j,l-1}$

Each model, $m_{j,l-1} \in M_{j,l-1}$, is defined as a triple, $m_{j,l-1} = \langle h_j, f_j, O_j \rangle$, where $f_j : H_j \to \Delta(A_j)$ is agent $j$'s function, assumed computable, which maps possible histories of $j$'s observations, $H_j$, to distributions over its actions. $h_j$ is an element of $H_j$, and $O_j$ is a function, also computable, specifying the way the environment is supplying the agent with its input. For simplicity, we may write model $m_{j,l-1}$ as $m_{j,l-1} = \langle h_j, \widehat{m}_j \rangle$, where $\widehat{m}_j$ consists of $f_j$ and $O_j$.

A specific class of models are the $(l-1)^{th}$ level *intentional* models, $\Theta_{j,l-1}$, of agent $j$: $\theta_{j,l-1} = \langle b_{j,l-1}, A, \Omega_j, T_j, O_j, R_j, OC_j \rangle$. $b_{j,l-1}$ is agent $j$'s belief nested to the level $l-1$, $b_{j,l-1} \in \Delta(IS_{j,l-1})$, and $OC_j$ is $j$'s optimality criterion. Rest of the notation is standard. We may rewrite $\theta_{j,l-1}$ as, $\theta_{j,l-1} = \langle b_{j,l-1}, \widehat{\theta}_j \rangle$, where $\widehat{\theta}_j \in \widehat{\Theta}_j$ includes all elements of the intentional model other than the belief and is called the agent $j$'s *frame*. The intentional models are analogous to *types* as used in Bayesian games (Harsanyi, 1967).

As mentioned by Gmytrasiewicz and Doshi (2005), we may also ascribe the *subintentional* models, $SM_j$, which constitute the remaining models in $M_{j,l-1}$. Examples of subintentional models are finite state controllers and fictitious play models (Fudenberg & Levine, 1998). While we do not consider these models here, they could be accommodated in a straightforward manner.

In order to promote understanding, let us define the finitely nested interactive state space in an inductive manner:

$$
\begin{array}{llll}
IS_{i,0} & = & S, & \Theta_{j,0} & = & \{\langle b_{j,0}, \widehat{\theta}_j \rangle : b_{j,0} \in \Delta(IS_{j,0}), A = A_j\}, \\
IS_{i,1} & = & S \times \Theta_{j,0}, & \Theta_{j,1} & = & \{\langle b_{j,1}, \widehat{\theta}_j \rangle : b_{j,1} \in \Delta(IS_{j,1})\}, \\
\cdot & & & \cdot \\
\cdot & & & \cdot \\
\cdot & & & \cdot \\
IS_{i,l} & = & S \times \Theta_{j,l-1}, & \Theta_{j,l} & = & \{\langle b_{j,l}, \widehat{\theta}_j \rangle : b_{j,l} \in \Delta(IS_{j,l})\}.
\end{array}
$$

Recursive characterizations of state spaces analogous to above have appeared previously in the game-theoretic literature (Mertens & Zamir, 1985; Brandenburger & Dekel, 1993; Battigalli & Siniscalchi, 1999) where they have led to the definitions of *hierarchical belief systems*. These have been proposed as mathematical formalizations of type spaces in Bayesian games. Additionally, the nested beliefs are, in general, analogous to hierarchical priors utilized for Bayesian analysis of hierarchical data (Gelman et al., 2004). Hierarchical priors arise when unknown priors are assumed to be drawn from a population distribution, whose parameters may themselves be unknown thereby motivating a higher level prior.

- $A = A_i \times A_j$ is the set of joint moves of all agents.

- $T_i$ is a transition function, $T_i : S \times A \times S \to [0, 1]$ which describes the results of the agent's actions on the physical states of the world. (It is assumed that actions can directly change the physical state only, see Gmytrasiewicz & Doshi, 2005).

- $\Omega_i$ is the set of agent $i$'s observations.

- $O_i$ is an observation function, $O_i : S \times A \times \Omega_i \to [0, 1]$ which gives the likelihood of perceiving observations in the state resulting from performing the action. (It is assumed that only the physical state is directly observable, and not the models of the other agent.)

- $R_i$ is defined as, $R_i : IS_i \times A \to \mathbf{R}$. While an agent is allowed to have preferences over physical states and models of other agents, usually only the physical state will matter.

### 4.1 Belief Update

Analogous to POMDPs, an agent within the I-POMDP framework also updates its belief as it acts and observes. However, there are two differences that complicate a belief update in multiagent settings, when compared to single agent ones. First, since the state of the physical environment depends on the actions performed by both agents, the prediction of how the physical state changes has to be made based on the predicted actions of the other agent. The probabilities of other's actions are obtained based on its models. Second, changes in the models of the other agent have to be included in the update. Specifically, since the other agent's model is intentional the update of the other agent's beliefs due to its new observation has to be included. In other words, the agent has to update its beliefs based on what it anticipates that the other agent observes and how it updates. The belief update function for an agent in the finitely nested I-POMDP framework is:

$$
\begin{aligned}
b_i^t(is^t) = \quad & \alpha \int_{is^{t-1}:\widehat{\theta}_j^{t-1}=\widehat{\theta}_j^t} b_{i,l}^{t-1}(is^{t-1}) \sum_{a_j^{t-1}} Pr(a_j^{t-1}|\theta_{j,l-1}^{t-1}) \, O_i(s^t, a^{t-1}, o_i^t) \, T_i(s^{t-1}, a^{t-1}, s^t) \\
& \times \sum_{o_j^t} \delta_D(SE_{\widehat{\theta}_j^t}(b_{j,l-1}^{t-1}, a_j^{t-1}, o_j^t) - b_{j,l-1}^t) \, O_j(s^t, a^{t-1}, o_j^t) \, d\, is^{t-1}
\end{aligned}
\tag{3}
$$

where $\alpha$ is the normalization constant, $\delta_D$ is the Dirac-delta function, $SE_{\widehat{\theta}_j^t}(\cdot)$ is an abbreviation denoting the belief update, and $Pr(a_j^{t-1}|\theta_{j,l-1}^{t-1})$ is the probability that $a_j^{t-1}$ is Bayes rational for the agent described by $\theta_{j,l-1}^{t-1}$.

If $j$ is also modeled as an I-POMDP, then $i$'s belief update invokes $j$'s belief update (via the term $SE_{\widehat{\theta}_j^t}(b_{j,l-1}^{t-1}, a_j^{t-1}, o_j^t)$), which in turn invokes $i$'s belief update and so on. This recursion in belief nesting bottoms out at the $0^{th}$ level. At this level, belief update of the agent reduces to a POMDP based belief update. [2] For an illustration of the belief update, additional details on I-POMDPs, and how they compare with other multiagent planning frameworks, see (Gmytrasiewicz & Doshi, 2005).

In a manner similar to the belief update in POMDPs, the following proposition holds for the I-POMDP belief update. The proposition results from noting that Eq. 3 expresses the belief in terms of parameters of the previous time step only. A complete proof of the belief update and this proposition is given by Gmytrasiewicz and Doshi (2005).

**Proposition 1. (Sufficiency)** *In a finitely nested I-POMDP$_{i,l}$ of agent $i$, $i$'s current belief, i.e., the probability distribution over the set $S \times \Theta_{j,l-1}$, is a sufficient statistic for the past history of $i$'s observations.*

### 4.2 Value Iteration

Each level $l$ belief state in I-POMDP$_{i,l}$ has an associated value reflecting the maximum payoff the agent can expect in this belief state:

$$
\begin{aligned}
U^t(\langle b_{i,l}, \widehat{\theta}_i \rangle) = \max_{a_i \in A_i} \Big\{ \quad & \int_{is \in IS_{i,l}} ER_i(is, a_i) b_{i,l}(is) d\, is + \\
& \gamma \sum_{o_i \in \Omega_i} Pr(o_i|a_i, b_{i,l}) U^{t-1}(\langle SE_{\widehat{\theta}_i}(b_{i,l}, a_i, o_i), \widehat{\theta}_i \rangle) \Big\}
\end{aligned}
\tag{4}
$$

---

2. The $0^{th}$ level model is a POMDP: Other agent's actions are treated as exogenous events and folded into T, O, and R.

where, $ER_i(is, a_i) = \sum_{a_j} R_i(is, a_i, a_j)Pr(a_j|\theta_{j,l-1})$ (since $is = (s, \theta_{j,l-1})$).

Eq. 4 is a basis for value iteration in I-POMDPs, and can be succinctly rewritten as $U^t = HU^{t-1}$, where $H$ is commonly known as the *value backup* operator. Analogous to POMDPs, $H$ is both isotonic and contracting, thereby making the value iteration convergent (Gmytrasiewicz & Doshi, 2005).

Agent $i$'s optimal action, $a_i^*$, for the case of finite horizon with discounting, is an element of the set of optimal actions for the belief state, $OPT(\theta_i)$, defined as:

$$OPT(\langle b_{i,l}, \widehat{\theta}_i\rangle) = \underset{a_i \in A_i}{argmax} \left\{ \int_{is \in IS_{i,l}} ER_i(is, a_i)b_{i,l}(is)d\,is + \gamma \sum_{o_i \in \Omega_i} Pr(o_i|a_i, b_{i,l})U(\langle SE_{\widehat{\theta}_i}(b_{i,l}, a_i, o_i), \widehat{\theta}_i\rangle) \right\}$$

(5)

## 5. Example: The Multiagent Tiger Problem

To illustrate our approximation methods, we utilize the multiagent tiger problem as an example. The multiagent tiger problem is a generalization of the single agent tiger problem outlined in Section 3 to the multiagent setting. For the sake of simplicity, we restrict ourselves to a two-agent setting, but the problem is extensible to more agents in a straightforward way.

In the two-agent tiger problem, each agent may open doors or listen. To make the interaction more interesting, in addition to the usual observation of growls, we added an observation of door creaks, which depends on the action executed by the other agent. Creak right (CR) is likely due to the other agent having opened the right door, and similarly for creak left (CL). Silence (S) is a good indication that the other agent did not open doors and listened instead. We assume that the accuracy of creaks is 90%, while the accuracy of growls is 85% as before. Again, the tiger location is chosen randomly in the next time step if any of the agents opened any doors in the current step. We also assume that the agent's payoffs are analogous to the single agent version. Note that the result of this assumption is that the other agent's actions do not impact the original agent's payoffs directly, but rather indirectly by resulting in states that matter to the original agent. Table 1 quantifies these factors.

When an agent makes its choice in the multiagent tiger problem, it may find it useful to consider what it believes about the location of the tiger, as well as whether the other agent will listen or open a door, which in turn depends on the other agent's beliefs, preferences and capabilities. In particular, if the other agent were to open any of the doors, the tiger's location in the next time step would be chosen randomly. The information that the agent had about the tiger's location till then, would reduce to zero. We simplify the situation somewhat by assuming that all of the agent $j$'s properties, except for beliefs, are known to $i$, and that $j$'s time horizon is equal to $i$'s. In other words, $i$'s uncertainty pertains only to $j$'s beliefs and not to its frame.

## 6. Representing Prior Nested Beliefs

As we mentioned, there is an infinity of intentional models of an agent. Since an agent is unaware of the true models of interacting agents *ex ante*, it must maintain a belief over all possible candidate models. The complexity of this space precludes practical implementations of I-POMDPs for all

| ⟨$a_i, a_j$⟩ | State | TL | TR |
|---|---|---|---|
| ⟨OL,*⟩ | * | 0.5 | 0.5 |
| ⟨OR,*⟩ | * | 0.5 | 0.5 |
| ⟨*,OL⟩ | * | 0.5 | 0.5 |
| ⟨*,OR⟩ | * | 0.5 | 0.5 |
| ⟨L,L⟩ | TL | 1.0 | 0 |
| ⟨L,L⟩ | TR | 0 | 1.0 |

Transition function: $T_i = T_j$

| ⟨$a_i, a_j$⟩ | TL | TR |
|---|---|---|
| ⟨OR,OR⟩ | 10 | -100 |
| ⟨OL,OL⟩ | -100 | 10 |
| ⟨OR,OL⟩ | 10 | -100 |
| ⟨OL,OR⟩ | -100 | 10 |
| ⟨L,L⟩ | -1 | -1 |
| ⟨L,OR⟩ | -1 | -1 |
| ⟨OR,L⟩ | 10 | -100 |
| ⟨L,OL⟩ | -1 | -1 |
| ⟨OL,L⟩ | -100 | 10 |

| ⟨$a_i, a_j$⟩ | TL | TR |
|---|---|---|
| ⟨OR,OR⟩ | 10 | -100 |
| ⟨OL,OL⟩ | -100 | 10 |
| ⟨OR,OL⟩ | -100 | 10 |
| ⟨OL,OR⟩ | 10 | -100 |
| ⟨L,L⟩ | -1 | -1 |
| ⟨L,OR⟩ | 10 | -100 |
| ⟨OR,L⟩ | -1 | -1 |
| ⟨L,OL⟩ | -100 | 10 |
| ⟨OL,L⟩ | -1 | -1 |

Reward functions of agents $i$ and $j$

| ⟨$a_i, a_j$⟩ | State | ⟨ GL, CL ⟩ | ⟨ GL, CR ⟩ | ⟨ GL, S ⟩ | ⟨ GR, CL ⟩ | ⟨ GR, CR ⟩ | ⟨ GR, S ⟩ |
|---|---|---|---|---|---|---|---|
| ⟨L,L⟩ | TL | 0.85*0.05 | 0.85*0.05 | 0.85*0.9 | 0.15*0.05 | 0.15*0.05 | 0.15*0.9 |
| ⟨L,L⟩ | TR | 0.15*0.05 | 0.15*0.05 | 0.15*0.9 | 0.85*0.05 | 0.85*0.05 | 0.85*0.9 |
| ⟨L,OL⟩ | TL | 0.85*0.9 | 0.85*0.05 | 0.85*0.05 | 0.15*0.9 | 0.15*0.05 | 0.15*0.05 |
| ⟨L,OL⟩ | TR | 0.15*0.9 | 0.15*0.05 | 0.15*0.05 | 0.85*0.9 | 0.85*0.05 | 0.85*0.05 |
| ⟨L,OR⟩ | TL | 0.85*0.05 | 0.85*0.9 | 0.85*0.05 | 0.15*0.05 | 0.15*0.9 | 0.15*0.05 |
| ⟨L,OR⟩ | TR | 0.15*0.05 | 0.15*0.9 | 0.15*0.05 | 0.85*0.05 | 0.85*0.9 | 0.85*0.05 |
| ⟨OL,*⟩ | * | 1/6 | 1/6 | 1/6 | 1/6 | 1/6 | 1/6 |
| ⟨OR,*⟩ | * | 1/6 | 1/6 | 1/6 | 1/6 | 1/6 | 1/6 |

| ⟨$a_i, a_j$⟩ | State | ⟨ GL, CL ⟩ | ⟨ GL, CR ⟩ | ⟨ GL, S ⟩ | ⟨ GR, CL ⟩ | ⟨ GR, CR ⟩ | ⟨ GR, S ⟩ |
|---|---|---|---|---|---|---|---|
| ⟨L,L⟩ | TL | 0.85*0.05 | 0.85*0.05 | 0.85*0.9 | 0.15*0.05 | 0.15*0.05 | 0.15*0.9 |
| ⟨L,L⟩ | TR | 0.15*0.05 | 0.15*0.05 | 0.15*0.9 | 0.85*0.05 | 0.85*0.05 | 0.85*0.9 |
| ⟨OL,L⟩ | TL | 0.85*0.9 | 0.85*0.05 | 0.85*0.05 | 0.15*0.9 | 0.15*0.05 | 0.15*0.05 |
| ⟨OL,L⟩ | TR | 0.15*0.9 | 0.15*0.05 | 0.15*0.05 | 0.85*0.9 | 0.85*0.05 | 0.85*0.05 |
| ⟨OR,L⟩ | TL | 0.85*0.05 | 0.85*0.9 | 0.85*0.05 | 0.15*0.05 | 0.15*0.9 | 0.15*0.05 |
| ⟨OR,L⟩ | TR | 0.15*0.05 | 0.15*0.9 | 0.15*0.05 | 0.85*0.05 | 0.85*0.9 | 0.85*0.05 |
| ⟨*,OL⟩ | * | 1/6 | 1/6 | 1/6 | 1/6 | 1/6 | 1/6 |
| ⟨*,OR⟩ | * | 1/6 | 1/6 | 1/6 | 1/6 | 1/6 | 1/6 |

Observation functions of agents $i$ and $j$.

Table 1: Transition, reward, and observation functions for the multiagent tiger problem.

but the simplest settings. Approximations based on sampling use a finite set of sample points to represent a complete belief state.

In order to sample from nested beliefs we first need to represent them. Agent $i$'s level 0 belief, $b_{i,0} \in \Delta(S)$, is a vector of probabilities over each physical state: $b_{i,0} \stackrel{def}{=} \langle\ p_{i,0}(s_1),\ p_{i,0}(s_2),\ \ldots, p_{i,0}(s_{|S|})\ \rangle$. The first and second subscripts of $b_{i,0}$ denote the agent and the level of nesting, respectively. Since belief is a probability distribution, $\sum_{q=1}^{|S|} p_{i,0}(s_q) = 1$. We refer to this constraint as the simplex constraint. As we may write, $p_{i,0}(s_{|S|}) = 1 - \sum_{q=1}^{|S|-1} p_{i,0}(s_q)$, subsequently, only $|S| - 1$ probabilities are needed to specify a level 0 belief.

For the tiger problem, let $s_1 = TL$ and $s_2 = TR$. An example level 0 belief of $i$ for the tiger problem, $b_{i,0} \stackrel{def}{=} \langle p_{i,0}(TL), p_{i,0}(TR) \rangle$, is $\langle 0.7, 0.3 \rangle$ that assigns a probability of 0.7 to $TL$ and 0.3 to $TR$. Knowing $p_{i,0}(TL)$ is sufficient for a complete specification of the level 0 belief.

Agent $i$'s first level belief, $b_{i,1} \in \Delta(S \times \Theta_{j,0})$, is a vector of densities over $j$'s level 0 beliefs, one for each combination of state and $j$'s frame and possibly distinct from each other. Hence, there are $|S||\widehat{\Theta}_j|$ such densities: $b_{i,1} \stackrel{def}{=} \langle\; p_{i,1}^{\langle s,\widehat{\theta}_j\rangle_1}, \; p_{i,1}^{\langle s,\widehat{\theta}_j\rangle_2}, \; \ldots, \; p_{i,1}^{\langle s,\widehat{\theta}_j\rangle_{|S||\widehat{\Theta}_j|}} \;\rangle$, where $\langle s, \widehat{\theta}_j\rangle_k$, $k = 1, \ldots, |S||\widehat{\Theta}_j|$ is a particular state and $j$'s frame combination. Because of the simplex constraint, the sum of integrals of the level 1 densities over the level 0 beliefs must be 1. We observe that the level 1 densities may be represented using any family of probability distributions such as exponential family (Dobson, 2002) or polynomials that allow approximation of any function up to arbitrary accuracy. As such, the densities could exhibit any shape given that they satisfy the simplex constraint.

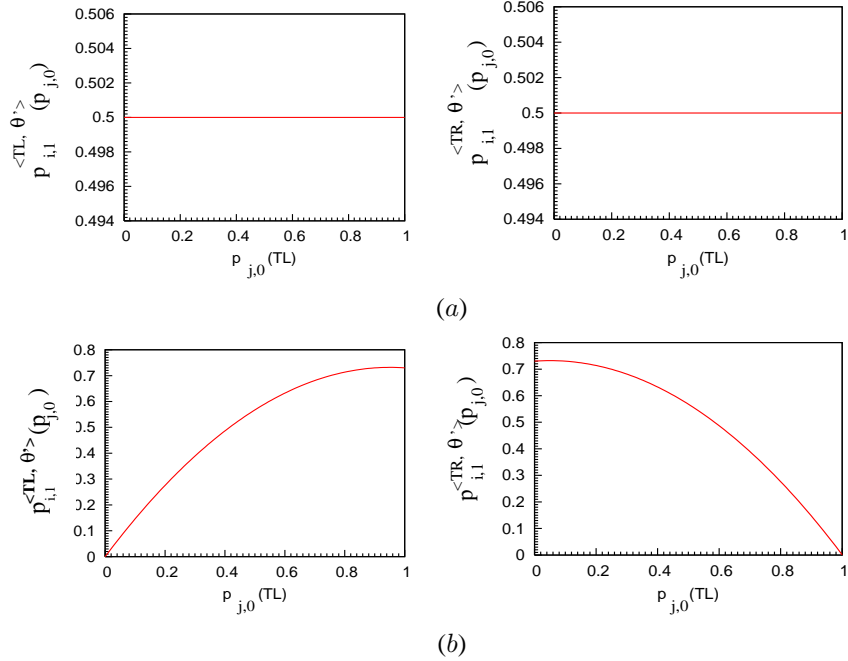

Figure 3: Example level 1 beliefs of $i$ in the two-agent tiger problem. ($a$) According to this belief, $i$ is uninformed of $j$'s level 0 beliefs. Since the marginal of each plot is 0.5, $i$ is also unaware of the location of the tiger. ($b$) Agent $i$ believes that $j$ likely knows the location of the tiger ($\int_0^{0.5} p_{i,1}^{\langle TL,\widehat{\theta}'\rangle}(p_{j,0})dp_{j,0} \ll \int_{0.5}^1 p_{i,1}^{\langle TL,\widehat{\theta}'\rangle}(p_{j,0})dp_{j,0}$), though $i$ itself is unaware of it as the marginal of each plot is 0.5.

An example level 1 belief of $i$, $b_{i,1} \stackrel{def}{=} \langle p_{i,1}^{\langle TL,\widehat{\theta}_j'\rangle}, p_{i,1}^{\langle TR,\widehat{\theta}_j'\rangle}\rangle$, in the tiger problem is one according to which $i$ is uninformed about $j$'s level 0 beliefs and about the location of the tiger (see Fig. 3(a)). The superscript, $\widehat{\theta}_j'$, is agent $j$'s frame that is known to $i$. Another example level 1 belief of $i$ is one according to which it believes that $j$ likely knows the location of the tiger (Fig. 3(b)).

Agent $i$'s second level belief, $b_{i,2} \in \Delta(S \times \Theta_{j,1})$, is a vector of densities over $j$'s level 1 beliefs for each state and $j$'s intentional frame. In comparison to level 0 and level 1 beliefs, representing doubly-nested beliefs and beliefs with deeper nestings is not trivial. This is because these are distributions over density functions whose representations need not be finite. For example, let $j$'s singly-nested belief densities be represented using the family of polynomials. Then, $i$'s doubly-nested belief over $j$'s densities is a vector of normalized mathematical functions of variables where the variables are the parameters of lower-level densities. Because the lower level densities are polynomials which could be of *any* degree and therefore any number of coefficients, the functions that represent doubly-nested beliefs may have an indefinite number of variables. Thus computable representations of $i$'s level 2 beliefs are not trivially obtained. We formalize this observation using Proposition 2, which shows that multiply-nested beliefs are necessarily partial functions that fail to assign a probability to some elements (lower level beliefs) in their domain.

**Proposition 2.** *Agent $i$'s multiply nested belief, $b_{i,l}$, $l \geq 2$, is strictly a partial recursive function.*

*Proof.* We briefly revisit the definition of nested beliefs: $b_{i,l} \in \Delta(IS_{i,l}) = \Delta(S \times \Theta_{j,l-1}) = \Delta(S \times \langle B_{j,l-1}, \widehat{\Theta}_j \rangle)$, where $B_{j,l-1}$ is the level $l - 1$ belief simplex and $\widehat{\Theta}_j$ is the set of frames of $j$. As the *basis* case, let $l = 2$, then $b_{i,2} \in \Delta(S \times \langle B_{j,1}, \widehat{\Theta}_j \rangle)$. Because the state and frame spaces are discrete, $b_{i,2}$ may be represented using a collection of density functions on $j$'s beliefs, one for each discrete state and $j$'s frame combination, $p_{i,2}^{\langle s,\widehat{\theta}_j \rangle}(b_{j,1})$, where $b_{j,1} \in B_{j,1}$. Notice that, $b_{j,1}$ being singly-nested, is itself a collection of densities over $j$'s level 0 beliefs, one for each state and $i$'s frame combination. Thus, as mentioned before let $b_{j,1} \stackrel{def}{=} \langle p_{j,1}^{\langle s,\widehat{\theta}_i \rangle_1}, p_{j,1}^{\langle s,\widehat{\theta}_i \rangle_2}, ..., p_{j,1}^{\langle s,\widehat{\theta}_i \rangle_{|S||\widehat{\Theta}_i|}} \rangle$.

Recall from Section 4 that the models and therefore the belief density functions are assumed computable. Let $x$ be the program of length in bits, $l(x)$, that encodes, say $p_{j,1}^{\langle s,\widehat{\theta}_i \rangle_1}$, in the language $g$. Then define the *complexity* of the density function, $p_{j,1}^{\langle s,\widehat{\theta}_i \rangle_1}$, as: $C_g(p_{j,1}^{\langle s,\widehat{\theta}_i \rangle_1}) = \min \{l(x) : g(x) = p_{j,1}^{\langle s,\widehat{\theta}_i \rangle_1}\}$. $C_g(\cdot)$ is the minimum length program in language $g$ that computes the argument.[3] We observe that $l(x)$ is proportional to the number of parameters that describe $p_{j,1}^{\langle s,\widehat{\theta}_i \rangle_1}$. Because the number of parameters of a density need not be bounded, $l(x)$ and consequently the complexity of the density may not be finite. Intuitively, this is equivalent to saying that the density could have "any shape".

Assume, by way of contradiction, that the level 2 density function, $p_{i,2}^{\langle s,\widehat{\theta}_j \rangle}$ is a total recursive function. Construct a Turing machine, $T$, that computes it. Because $p_{i,2}^{\langle s,\widehat{\theta}_j \rangle}$ is total, $T$ will halt on all inputs. Specifically, $T$ will read the set of symbols on its input tape that describe the level 1 density function (the program, $x$), and once it has finished reading it halts and leaves a number between 0 and 1 on the output tape. This number is the output of the density function encoded by $T$. Note that $T$ does not execute the input program $x$, but simply parses it to enable identification. Thus $T$ is not a universal Turing machine. As we mentioned previously, the minimum length program (and hence the complexity) that encodes the level $l$ density function may be infinite. Thus the size of the set of symbols on the input tape of $T$, $l(x)$, may be infinite, and $T$ may not halt. But this is a contradiction. Thus, $p_{i,2}^{\langle s,\widehat{\theta}_j \rangle}$ is a partial recursive function.

The argument may be extended inductively to further levels of nesting. □

As multiply-nested beliefs in their general form are partial recursive functions that are not defined for every possible lower level belief in their domain, restrictions on the complexity of the nested beliefs (where complexity is as defined in Proposition 2) are needed to allow for computability and so that they are well-defined. One way is to focus our attention on limited representations involving a bounded number of parameters.

---

3. Note that the complexity, $C_g$, is within a constant of the Kolmogorov complexity (Li & Vitanyi, 1997) of the density function, $p_{j,1}^{\langle s,\widehat{\theta}_i \rangle_1}$.

Because of these complications, a general-purpose language for representing nested beliefs is beyond the scope of this article and we do not attempt it here; it is a topic of continuing investigations. Instead, we utilize specific examples of doubly-nested and more deeply nested beliefs for our experiments in the remainder of this article.

## 7. Decomposing the I-POMDP Belief Update

Analogous to Eqs. 1 and 2 of the Bayes filter, we decompose the I-POMDP belief update into two steps. The decomposition not only facilitates a better understanding of the belief update, but also plays a pivotal role in the development of the approximation.

- *Prediction:* When an agent, say $i$, performs an action $a_i^{t-1}$, and agent $j$ performs $a_j^{t-1}$, the predicted belief state is:

$$
\begin{aligned}
Pr(is^t | a_i^{t-1}, a_j^{t-1}, b_{i,l}^{t-1}) =\ & \int_{IS^{t-1}: \widehat{\theta}_j^{t-1} = \widehat{\theta}_i^t}\ b_{i,l}^{t-1}(is^{t-1}) Pr(a_j^{t-1} | \theta_{j,l-1}^{t-1}) \\
& \times T_i(s^{t-1}, a_i^{t-1}, a_j^{t-1}, s^t) \sum_{o_j^t} O_j(s^t, a_i^{t-1}, a_j^{t-1}, o_j^t) \\
& \times \delta_D(SE_{\widehat{\theta}_j^t}(b_{j,l-1}^{t-1}, a_j^{t-1}, o_j^t) - b_{j,l-1}^t)\ d\, is^{t-1}
\end{aligned}
\tag{6}
$$

where $\delta_D$ is the Dirac-delta function, $SE_{\widehat{\theta}_j^t}(\cdot)$ is an abbreviation for the belief update, $Pr(a_j^{t-1} | \theta_{j,l-1}^{t-1})$ is the probability that $a_j^{t-1}$ is Bayes rational for the agent described by $\theta_{j,l-1}^{t-1}$.

- *Correction:* When agent $i$ perceives an observation, $o_i^t$, the corrected belief state is a weighted sum of the predicted belief states for each possible action of $j$:

$$
Pr(is^t | o_i^t, a_i^{t-1}, b_{i,l}^{t-1}) = \alpha \sum_{a_j^{t-1}} O_i(s^t, a_i^{t-1}, a_j^{t-1}, o_i^t) Pr(is^t | a_i^{t-1}, a_j^{t-1}, b_{i,l}^{t-1})
\tag{7}
$$

where $\alpha$ is the normalizing constant.

Equations 6 and 7 along with Proposition 1 may be seen as a generalization of the single agent Bayes filter (see Section 3) to the multiagent setting. At a general level, these equations represent an application of the update of hierarchical priors given observed data (Gelman et al., 2004) to the problem of state estimation in multiagent settings.

## 8. Interactive Particle Filter for the Multiagent Setting

We presented the algorithm for the traditional bootstrap filter in Section 3. As we mentioned before, the bootstrap filter is a MC sampling based randomized implementation of the POMDP belief update (Bayes filter). We generalize this implementation so as to approximate the I-POMDP belief update steps presented previously in Section 7.

## 8.1 Description

Our generalization of the PF to the multiagent case, which we call an *interactive particle filter* (I-PF), similar to the basic PF, involves the key steps of *importance sampling* and *selection*. The resulting algorithm inherits the convergence properties of the original algorithm (Doucet et al., 2001). Specifically, the approximate posterior belief generated by the filter converges to the truth (as computed by Eqs. 6 and 7) as the number of particles ($N$) tends to infinity. Note that the presence of other agents does not affect the convergence because, ($a$) the exact belief update that provides the stationary point similarly reasons about the presence of other agents, and ($b$) we explicitly model other agents' actions due to which the nonstationarity in the environment vanishes.

The extension of the PF to the multiagent setting turns out to be nontrivial because we are faced with predicting the other agent's action(s), which requires us to deal with an interactive belief hierarchy. Analogously to the I-POMDP belief update, the I-PF reduces to the traditional PF when there is only one agent in the environment.

The I-PF, described in Fig. 4, requires an initial set of $N$ particles, $\tilde{b}_{k,l}^{t-1}$, that is approximately representative of the agent's prior belief, along with the action, $a_k^{t-1}$, the observation, $o_k^t$, and the level of belief nesting, $l > 0$. As per our convention, $k$ will stand for either agent $i$ or $j$, and $-k$ for the other agent, $j$ or $i$, as appropriate. Each particle, $is_k^{(n)}$, in the sample set represents the agent's possible interactive state, in which the other agent's belief may itself be a set of particles. Formally, $is_k^{(n)} = \langle s^{(n)}, \theta_{-k}^{(n)} \rangle$, where $\theta_{-k}^{(n)} = \langle \tilde{b}_{-k,l-1}^{(n)}, \hat{\theta}_{-k}^{(n)} \rangle$. Note that $\tilde{b}_{k,0}^{(n)}$ is a probability distribution over the physical state space.

We generate $\tilde{b}_{k,l}^{t-1}$ by sampling $N$ particles from the prior nested belief. Given a prior nested belief, this is a simple recursive procedure that first uses the marginals over the physical states and frames at the current nesting level to sample the state and frame of the other agent. We then sample $N$ particles from the density over the lower level beliefs, conditioned on the sampled state and frame combination. If the belief is multiply nested, this operation is recursively performed bottoming out at the lowest level where the other agent's flat (level 0) beliefs are sampled.

The interactive particle filtering proceeds by *propagating* each particle forward in time. However, as opposed to the traditional particle filtering, *this is not a one-step process*: ($i$) In order to perform the propagation, other agent's action must be known. As the model ascribed to the other agent is intentional, this is obtained by solving the other agent's model (using the algorithm AP-PROXPOLICY described later in Section 9) to find a distribution over its actions, from which its action is sampled (lines 3–4 in Fig. 4). Specifically, if OPT is the set of optimal actions obtained by solving the model, then $Pr(a_{-k}) = \frac{1}{|OPT|}$ if $a_{-k}$ is in OPT, 0 otherwise. ($ii$) Additionally, analogous to the exact belief update, for each of the other agent's possible observations, we must update its model (line 6). Because the model is intentional, we must update its belief state. If $l > 1$, updating the other agent's belief requires recursively invoking the I-PF for performing its belief update (lines 12–14). This recursion in depth of the belief nesting terminates when the level of nesting becomes one, and a LEVEL0BELIEFUPDATE, described in Fig. 5, is performed (lines 8–10).[4] In addition to using the agent's own observation for *weighting*, the other agent's observations also participate in the weighting process (lines 15–16). The latter allows us to distinguish $j$'s beliefs that are more likely given the physical state. Though the propagation and weighting steps generate $|\Omega_{-k}|N$ appropriately weighted particles, we *resample* $N$ particles out of these (line 19), using an unbiased

---

4. If the physical state space is also continuous or very large, then we would replace the level 0 belief update with a traditional particle filter. However, in doing so, we would loose the theoretical bounds given in Section 9.1

---

**Function** I-PARTICLEFILTER ($\widetilde{b}_{k,l}^{t-1}, a_k^{t-1}, o_k^t, l > 0$) **returns** $\widetilde{b}_{k,l}^t$

1. $\widetilde{b}_{k,l}^{tmp} \leftarrow \phi, \widetilde{b}_{k,l}^t \leftarrow \phi$

   *Importance Sampling*

2. **for all** $is_k^{(n),t-1} = \langle s^{(n),t-1}, \theta_{-k}^{(n),t-1} \rangle \in \widetilde{b}_{k,l}^{t-1}$ **do**

3.      $Pr(A_{-k}|\theta_{-k}^{(n),t-1}) \leftarrow$ APPROXPOLICY($\theta_{-k}^{(n),t-1}, l-1$)

4.      Sample $a_{-k}^{t-1} \sim Pr(A_{-k}|\theta_{-k}^{(n)})$      *// Sample other agent's action*

5.      Sample $s^{(n),t} \sim T_k(S^t|a_k^{t-1}, a_{-k}^{t-1}, s^{(n),t-1})$   *// Sample the physical state*

6.      **for all** $o_{-k}^t \in \Omega_{-k}$ **do**      *// Update other agent's belief*

7.         **if** ($l = 1$) **then**      *// I-POMDP is singly nested*

8.            $b_{-k,0}^{(n),t} \leftarrow$ LEVEL0BELIEFUPDATE($b_{-k,0}^{(n),t-1}, a_{-k}^{t-1}, o_{-k}^t$)

9.            $\theta_{-k}^{(n),t} \leftarrow \langle b_{-k,0}^{(n),t}, \widehat{\theta}_{-k}^{(n)} \rangle$

10.           $is_k^{(n),t} \leftarrow \langle s^{(n),t}, \theta_{-k}^{(n),t} \rangle$

11.         **else**      *// I-POMDP is multiply nested*

12.            $\widetilde{b}_{-k,l-1}^{(n),t} \leftarrow$ I-PARTICLEFILTER($\widetilde{b}_{-k,l-1}^{(n),t-1}, a_{-k}^{t-1}, o_{-k}^t, l-1$)

13.            $\theta_{-k}^{(n),t} \leftarrow \langle \widetilde{b}_{-k,l-1}^{(n),t}, \widehat{\theta}_{-k}^{(n)} \rangle$

14.           $is_k^{(n),t} \leftarrow \langle s^{(n),t}, \theta_{-k}^{(n),t} \rangle$

15.      Weight $is_k^{(n),t}$: $w_t^{(n)} \leftarrow O_{-k}(o_{-k}^t|s^{(n),t}, a_k^{t-1}, a_{-k}^{t-1})$

16.      Adjust weight: $w_t^{(n)} \leftarrow w_t^{(n)} \times O_k(o_k^t|s^{(n),t}, a_k^{t-1}, a_{-k}^{t-1})$

17.      $\widetilde{b}_{k,l}^{tmp} \overset{\cup}{\leftarrow} (is_k^{(n),t}, w_t^{(n)})$

18. Normalize all $w_t^{(n)}$ so that $\sum_{n=1}^N w_t^{(n)} = 1$

   *Selection*

19. Resample with replacement $N$ particles $\{is_k^{(n),t}, n = 1...N\}$
   from the set $\widetilde{b}_{k,l}^{tmp}$ according to the importance weights.

20. $\widetilde{b}_{k,l}^t \leftarrow \{is_k^{(n),t}, n = 1...N\}$

21. **return** $\widetilde{b}_{k,l}^t$

**end function**

---

Figure 4: Interactive particle filtering for approximating the I-POMDP belief update. A nesting of filters is used to update all levels of the belief. $k$ denotes either agent $i$ or $j$, and $-k$ the other agent, $j$ or $i$, as appropriate. Also see Fig. 6 for a visualization.

resampling scheme. Lines 2–15 represent a simulation of the prediction step (Eq. 6), while lines 16–20 are the simulated implementation of the correction step (Eq. 7).

An alternative approach within the propagation step is to sample the other agent's observation because it is a hidden variable. We may then update its belief given the sampled observation. Although statistically equivalent to our approach, it involves an additional step of sampling, which further contributes to the sources of error in the I-PF. In particular, for lesser number of particles, other agent's beliefs resulting from low probability observations may not appear in the resampled posterior. This is because other agent's low probability observations are less likely to be sampled. Because the original agent's observation is independent of the other's belief, particles with identical physical states but different beliefs of the other are weighted equally. As the beliefs resulting from low probability observations are less frequent in the sample set, they are less likely to be picked in the resampled posterior. In comparison, weighting using the other agent's observation removes

---

**Function** LEVEL0BELIEFUPDATE $(b_{k,0}^{t-1}, a_k^{t-1}, o_k^t)$ **returns** $b_{k,0}^t$
1.  $Pr(a_{-k}^{t-1}) \leftarrow 1/a_{-k}^{t-1}$            // *Other agent's action as noise*
2.  **for all** $s^t \in S$ **do**
3.      sum $\leftarrow 0$
4.      **for all** $s^{t-1} \in S$ **do**
5.         $Pr(s^t|s^{t-1}, a^{t-1}) \leftarrow 0$
6.         **for all** $a_{-k}^{t-1} \in A_{-k}$ **do**      // *Marginalize noise*
7.            $Pr(s^t|s^{t-1}, a_k^{t-1}) \overset{+}{\leftarrow} T_k(s^t|s^{t-1}, a_k^{t-1}, a_{-k}^{t-1})Pr(a_{-k}^{t-1})$
8.         sum $\overset{+}{\leftarrow} Pr(s^t|s^{t-1}, a_k^{t-1})b_k^{t-1}(s^{t-1})$
9.      $Pr(o_k^t|s^t, a_k^{t-1}) \leftarrow 0$
10.     **for all** $a_{-k}^{t-1} \in A_{-k}$ **do**        // *Marginalize noise*
11.        $Pr(o_k^t|s^t, a_k^{t-1}) \overset{+}{\leftarrow} O_k(o_k^t|s^t, a_k^{t-1}, a_{-k}^{t-1})Pr(a_{-k}^{t-1})$
12.     $b_{k,0}^t(s^t) \leftarrow Pr(o_k^t|s^t, a_k^{t-1})\times$ sum
13. Normalize the belief, $b_k^t$
14. **return** $b_k^t$
**end function**

Figure 5: The level 0 belief update which is similar to the exact POMDP belief update with the other agent's actions treated as noise. As an example, the noise may simply be a uniform distribution over the other agent's actions.

the intermediate sampling step and a source of error but at the expense of temporarily generating a larger number of particles. Based on a preference for reduced approximation error or computational efficiency, one of the two alternative steps may be used.

A visualization of our I-PF implementation is shown in Fig. 6. Note that the number of particles grows exponentially with the nesting level, due to which the approach becomes intractable for a larger number of levels. A method to limit the number of particles as we descend through the nesting level is needed to address this source of complexity. This is one line of our future work.

Notice that the I-PF could also be viewed as a recursive implementation of an approximately Rao-Blackwellised particle filter (RBPF) (Doucet, de Freitas, Murphy, & Russell, 2000), where the conditional distribution over models is itself updated using a RBPF. Doshi (2007) presented a Rao-Blackwellised I-PF (RB-IPF), where the conditional distribution is updated using a variational Kalman filter. Although the performance of the RB-IPF improves on the I-PF, it is restricted to prior beliefs that are Gaussian and could not be generalized to beyond a single level of nesting.

## 8.2 Illustration of the I-PF

We illustrate the operation of the I-PF using the multiagent tiger problem introduced in Section 5. For the sake of simplicity we consider $i$'s prior belief to be singly nested, i.e. $l = 1$. The procedure is recursively performed for more deeply nested beliefs. Let $i$ be uninformed about $j$'s level 0 beliefs, and about the location of the tiger (see Fig. 3$(a)$). We demonstrate the operation of the I-PF for the case when $i$ listens (L) and hears a growl from the left and no creaks, $\langle$GL,S$\rangle$.

In Fig. 7, we show the initial sample set, $\widetilde{b}_{i,1}^{t-1}$, consisting of $N = 2$ particles that is approximately representative of $i$'s singly-nested beliefs. Since we assume that $j$'s frame is known, each particle is an interactive state consisting of the tiger's location and $j$'s level 0 belief. Let a belief of 0.5 of $j$ be sampled from each of the flat line densities.

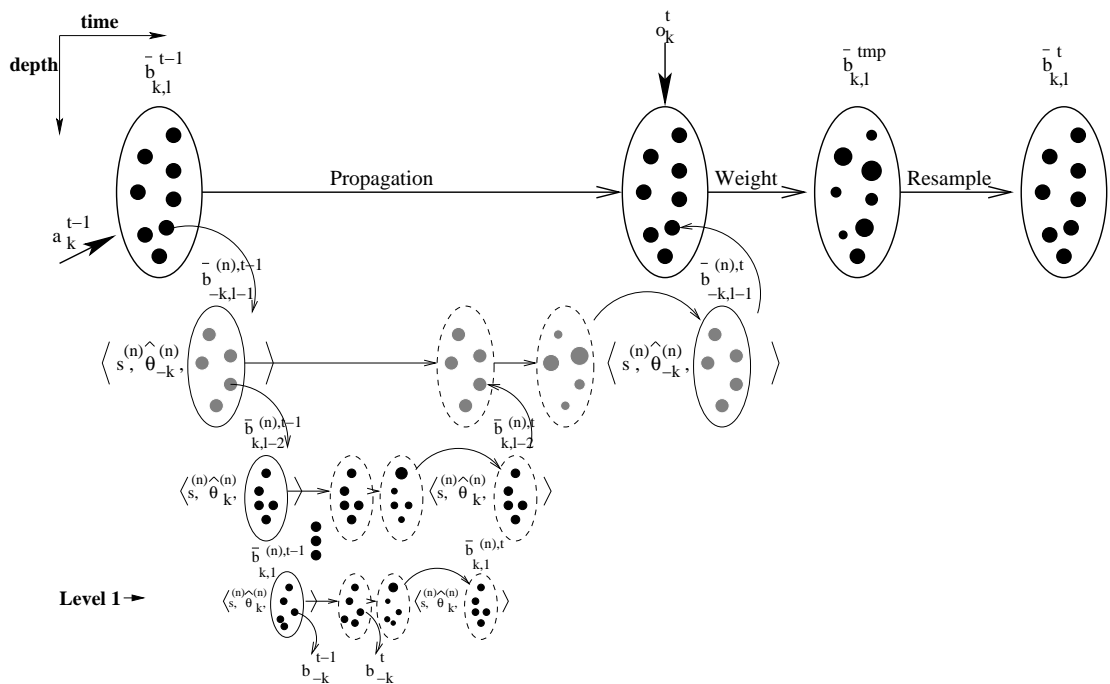

Figure 6: An illustration of the nesting in the I-PF. Colors black and gray distinguish filtering for the two agents. Because the propagation step involves updating the other agent's beliefs, we perform particle filtering on its beliefs. The filtering terminates when it reaches the level 1 nesting, where a level 0 belief update is performed for the other agent.

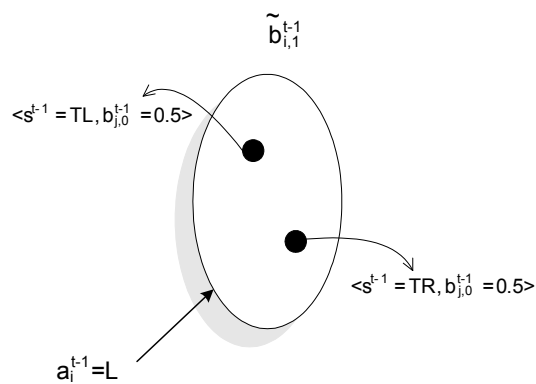

Figure 7: An initial sample set of 2 particles that is approximately representative of $b_{i,1}^{t-1}$ shown in Fig. 3(a).

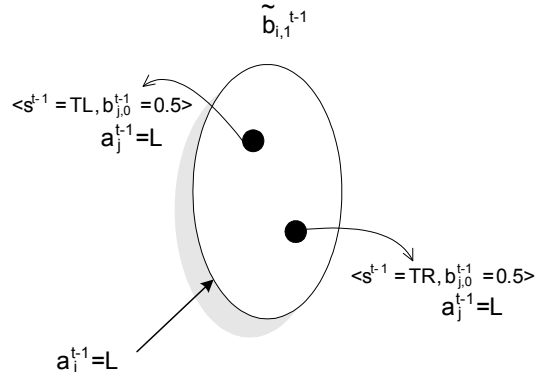

Figure 8: The initial sample set with $j$'s optimal action shown for each particle.

As we mentioned before, the *propagation* of the particles from time step $t-1$ to $t$ is a two-step process. As the first step, we solve $j$'s POMDP to compute its optimal action when its belief is $0.5$. $j$'s action is to listen since it does know the location of the tiger. We depict this in Fig. 8.

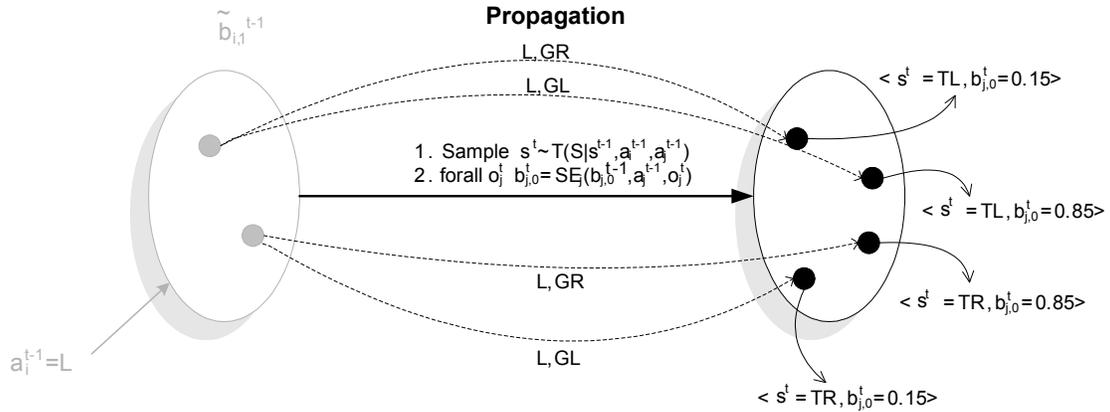

Figure 9: The propagation of the particles from time step $t-1$ to time step $t$. It involves sampling the next physical state and updating $j$'s beliefs by anticipating its observations (denoted using dashed arrows). Because $j$ may receive any one of two observations, there are 4 particles in the propagated sample set.

The second step of the propagation is to sample the next physical state for each particle using the transition function. Since both $i$ and $j$ listen, the location of the tiger remains unchanged. Next, we must update $j$'s beliefs. We do this by anticipating what $j$ might observe, and updating its belief exactly given its optimal action of listening. Since $j$ could receive one of two possible observations – GL or GR – each particle "splits" into two. This is shown using the dashed arrows going from particles in the initial sample set to the particles in the propagated sample set, in Fig. 9. When $j$ hears a GL, its updated belief is $0.85$ (that the tiger is behind the left door), otherwise it is $0.15$ when it hears a GR. If the level of belief nesting is greater than one, $j$'s belief update would be performed by recursively invoking the interactive particle filter on $j$'s beliefs.

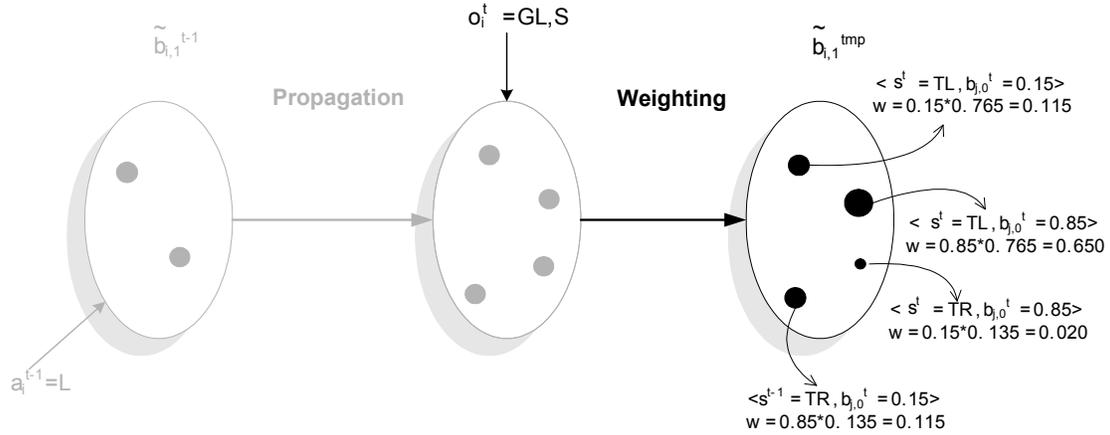

Figure 10: The weighting step is a two step process: Each particle is first weighted with the likelihood with which $j$ receives its observations, followed by adjusting this weight using the probability of $i$ making its observation of $\langle$GL,S$\rangle$. Note that resulting weights as shown are not normalized.

As part of the *weighting*, we will first weight each particle with the probability of $j$ receiving its observations. Particles with larger weights contain beliefs of $j$ that are more likely. Thereafter, we will scale this weight with the probability of $i$ observing a growl from the left and no creaks, $\langle$GL,S$\rangle$. To understand the weighting process, let's focus on a single particle. Weighting for the remaining particles is analogous.

We consider the particle on the top right in the sample set, $\widetilde{b}_{i,1}^{tmp}$, shown in Fig. 10. Agent $j$'s level 0 belief of 0.85 in this particle is due to $j$ hearing a growl from the left on listening. The probability of $j$ making this observation as given by its observation function, when the tiger is on the left is 0.85. We will adjust this weight with the probability of $i$ perceiving $\langle$GL,S$\rangle$ when the tiger is on the left and both agents are listening. This probability as given by $i$'s observation function is 0.765 (see Table 1). The final weight attached to this particle is 0.65. Note that the weights as shown in Fig. 10 are not normalized. After normalization $i$'s belief that the tiger is on the left is 0.85 (obtained by adding the normalized weights for particles that have $s^t$=TL), and 0.15 for tiger on the right. We note that this conforms to our expectations.

The final step of the I-PF is an unbiased resampling of the particles using the weights as the distribution. To prevent an exponential growth in the number of particles [5], we resample $N$ particles resulting in the sample set, $\widetilde{b}_{i,1}^{t}$, that is approximately representative of the exactly updated belief.

When $i$'s belief is nested to deeper levels, the above mentioned example forms the bottom step of the recursive filtering process.

### 8.3 Performance of the I-PF

As part of our empirical investigation of the performance of the I-PF, we show, using a standard pseudo-distance metric and visually, that it approximates the exact state estimation closely. We begin by utilizing extended versions of standard test problems and proceed to demonstrate the performance on a larger problem.

---

5. After $t$ propagation steps, there will be $N|\Omega_j|^t$ particles in the sample set.

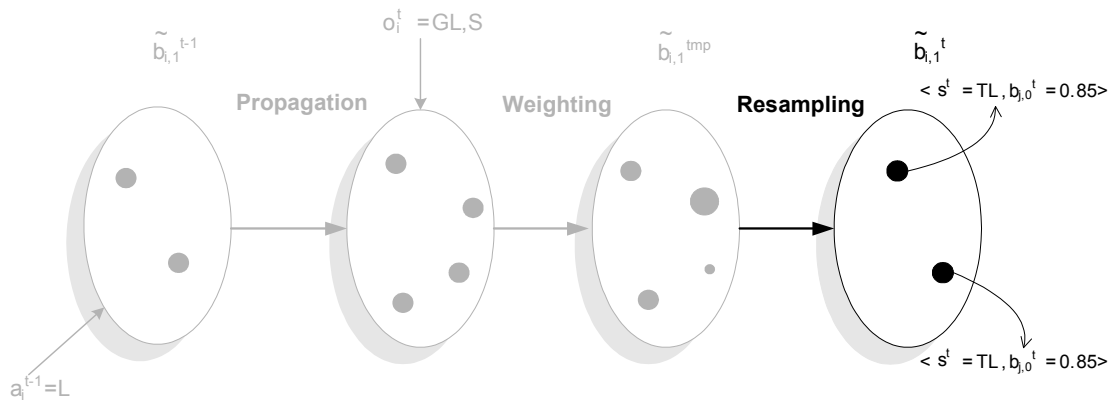

Figure 11: The final step is an unbiased resampling using the weights as the distribution.

### 8.3.1 MULTIAGENT TIGER AND MACHINE MAINTENANCE PROBLEMS

For our analysis, we first utilize the two-agent tiger problem that has two physical states, as described in Section 5, and a two-agent version of the machine maintenance problem (MM) (Smallwood & Sondik, 1973) described in detail in Appendix A, that has three physical states. While these problems have few physical states, the interactive state space tends to get large as it includes models of the other agent. Due to an absence of other general approximation techniques for I-POMDPs, we use a grid based numerical integration implementation of the exact filter as the baseline approximation for comparison. We obtained the points for numerical integration by superimposing regular grids of differing resolutions on the interactive state space.

The lineplots in Fig. 12 show that the quality of the I-PF based approximation, as measured by KL-Divergence becomes better as the number of particles increases, for both the problem domains. This remains true for both level 1 and 2 beliefs. KL-Divergence measures the difference between the two probability distributions by giving the relative entropy of the filtered posterior with respect to the near-exact one as obtained from the numerical integration. Note that the performance of the I-PF remains consistent over both the two-state tiger and the three-state MM problem. However, level 2 belief approximations require considerably more particles as compared to level 1 approximations, to achieve similar performance, indicating that the performance of the I-PF is affected by the level of nesting. Each data point in the lineplots is the average of 10 runs of the I-PF on multiple prior belief states. In the case of the tiger problem, the posterior used for comparison is the one that is obtained after agent $i$ listens and hears a growl from the left and no creaks. For the MM problem, the posterior obtained after $i$ manufactures and perceives no defect in the product, is used for comparison. Two of the prior level 1 beliefs of agent $i$ when playing the tiger problem are those shown in Fig. 3. We considered a level 2 belief according to which agent $i$ is unaware of the tiger's location and believes with equal probabilities that either of the level 1 beliefs shown in Fig. 3 are likely. We utilized analogous beliefs for the machine maintenance problem.

A comparison of the run times of the I-POMDP belief update implemented using grid based numerical integration and the I-PF is shown in Table. 2. We varied the number of grid points and the number of particles in the two implementations, respectively. As we use initial beliefs that are flat, having the same number of grid points and particles provides for a comparable approximation quality. The I-PF implementation significantly outperforms the numerical integration based implementation, while providing the comparable performance quality. Additionally, the run times of the grid based implementation increase significantly more when we move from the two-state tiger prob-lem to the three-state MM problem, in comparison to the increase for the I-PF for level 1 beliefs. Since the level 1 multiagent tiger model has 6 observations in comparison to 2 for the multiagent MM, the run times decrease as we move from the tiger to the MM problem for level 2 beliefs.

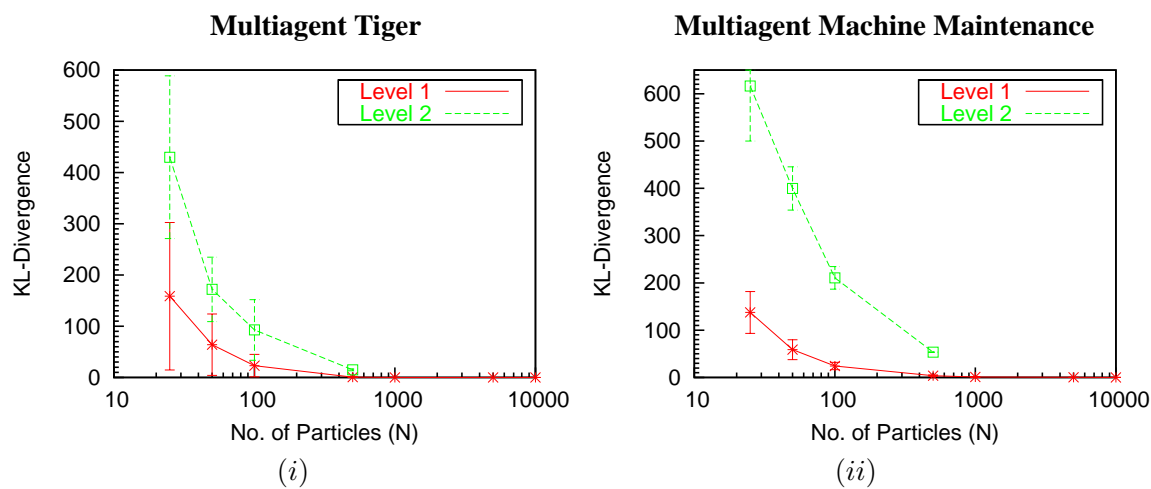

Figure 12: Anytime performance of the I-PF as a function of the number of particles, on $(i)$ multiagent tiger problem, $(ii)$ multiagent machine maintenance problem. The performance of the I-PF significantly improves with an increase in the number of particles leading toward convergence to the true posterior. The vertical bars are the standard deviations.

| Belief | Problem | Method | N=500 | N=1000 |
|---|---|---|---|---|
| Level 1 | Multiagent Tiger | I-PF | 0.148s ± 0.001s | 0.332s ± 0.007s |
| | | Grid based | 21.65s ± 0.18s | 1m 25.19s 1.02s |
| | Multiagent MM | I-PF | 0.452s ± 0.009s | 0.931s ± 0.0146s |
| | | Grid based | 1m 0.28s ± 0.21s | 3m 27.87s ± 0.13s |
| Level 2 | Multiagent Tiger | I-PF | 2m 23.28s ± 1.1s | 11m 41.30s ± 1.52s |
| | | Grid based | 34m 40.36s ± 0.55s | 77m 2.9s ± 4.88s |
| | Multiagent MM | I-PF | 1m 37.59s ± 0.17s | 8m 27.29s ± 1.65s |
| | | Grid based | 24m 55.33s ± 4.77s | 56m 21.97s ± 5.73s |

Table 2: Comparison of the average running times of our grid based numerical integration and I-PF implementations on the same platform (*Pentium IV, 1.7GHz, 512MB RAM, Linux*).

Despite using an equal number of grid points and particles, the reduced run time of the I-PF in comparison to the grid based approach is due to: $(i)$ iterating over all grid points in order to obtain the belief over each interactive state under consideration. In contrast, the I-PF iterates over all particles at each level once – propagating and weighting them to obtain the posterior; $(ii)$ the I-PF solves the models approximately in comparison to solving them exactly over the grid points; and $(iii)$ while the grid based belief update considers all the optimal actions for a model, the I-PF samples a single action from the distribution and propagates the corresponding particle using this action.

### Level 1 Beliefs in the Multiagent Tiger Problem

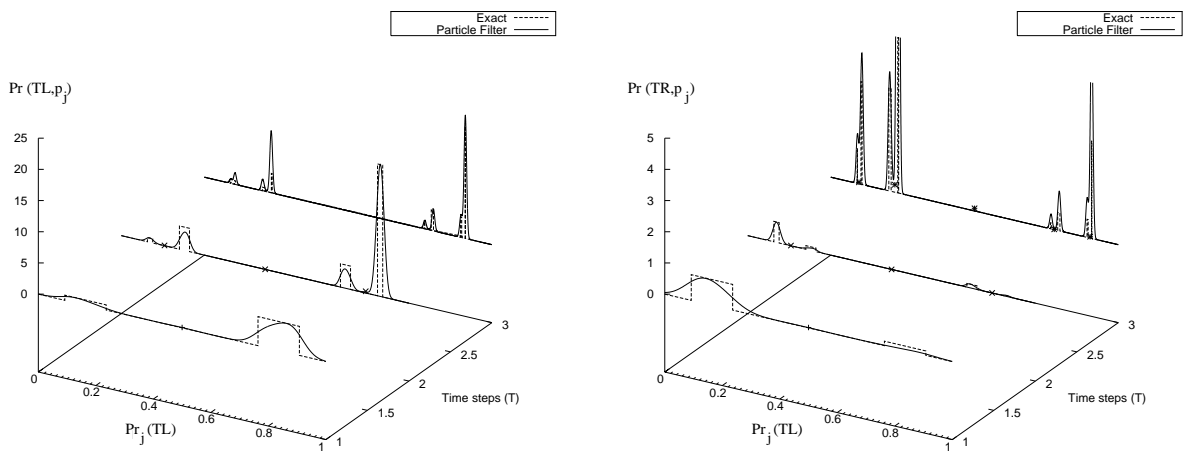

Figure 13: The exact and approximate p.d.f.s after successive filtering steps. The peaks of the approximate p.d.f.s align correctly with those of the exact p.d.f.s, and the areas under the approximate and exact p.d.f.s are approximately equal.

In order to assess the quality of the approximations after *successive* belief updates, we graphed the probability density functions produced by the I-PF and the exact belief update. The densities arising after each of three filtering steps on the level 1 belief of agent $i$ (Fig. 3(a)) in the tiger problem, are shown in Fig. 13. Each approximate p.d.f. is the average of 10 runs of the I-PF which contained 5000 particles, and is shaped using a standard Gaussian kernel. Gmytrasiewicz and Doshi (2005) provide an explanation of the exact I-POMDP belief update shown here. Briefly, as the prior belief is a flat line, the posterior becomes segmented where the segments correspond to beliefs of $j$ that are likely based on the predicted action and anticipated observations of $j$. The height of a segment is proportional to the likelihood of j's possible observation. The action and observation sequence followed was $\langle L, GL, S \rangle, \langle L, GL, S \rangle, \langle OR, GL, S \rangle$. As can be seen, the I-PF produces a good approximation of the true densities.

### 8.3.2 A UAV RECONNAISSANCE PROBLEM

Unmanned agents such as UAVs are finding important applications in tasks such as fighting forest fires (Casbeer, Beard, McLain, Sai-Ming, & Mehra, 2005; Sarris, 2001), law enforcement (Murphy & Cycon, 1998) and reconnaissance in warfare. The agents must operate in complex environments characterized by multiple parameters that affect their decisions, including, particularly in warfare, other agents who may have antagonistic preferences. The task is further complicated because the agent may possess noisy sensors and unreliable actuators.

We consider the task of a UAV $i$ which performs low-altitude reconnaissance of a potentially hostile theater that may be populated by other agents with conflicting objectives that serve as ground reconnaissance targets (see Fig. 14($a$)). To facilitate our analysis, we divide the UAV's operating theater into a $3 \times 3$ grid of sectors and consider a ground reconnaissance target ($T$), which could be located in any of the 9 sectors. For example, the target could be a terrorist who may be hiding in a safe house located in each sector. Of course, the UAV is unaware of which sector contains $T$, but is aware of its own location. To assist in its goal of spotting $T$ by moving to the sector containing it, the UAV may receive a noisy communication which informs it about the rows (similar colored sectors in Fig. 14) that likely contain $T$, though with some uncertainty. The UAV has the option of moving in any of the four cardinal directions to the adjacent sector, or hovering in its current location and listening for more communications.

The target $T$ may be informed (by its collaborators) that it could be in danger of being spotted by the UAV although with high uncertainty, in which case $T$ may move to an adjacent or diagonal sector. Note that the actions of both, the UAV and agent $T$ may affect the physical state of the problem. We formulate the decision problem of the UAV below:

• A physical state, $s=\{row_i, \; side_i \; or \; center_i, \; row_T, \; side_T \; or \; center_T\}$, where $row_i$ and $side_i \; or$ $center_i$ indicate the row location of UAV $i$ and whether the UAV is located in the side columns or the center column, respectively; • The joint action space, $A = A_i \times A_T$, where $A_i = \{move_N, \ldots, move_W,$ $listen\}$ and $A_T = \{move_N, \ldots, move_W, listen\}$. Here, $move_x$ moves the UAV or the target in the direction indicated by $x$, and $listen$ denotes the act of receiving communications about the location of $T$ or the UAV; • Observation space of the UAV is, $\Omega_i = \{$top-row (TR), center-row (CR), bottom-row (BR)$\}$, where for example, TR indicates that the corresponding target is in one of the three sectors in the top row; • Transition function is, $T_i : S \times A \times S \rightarrow [0,1]$. $T_i$ models the fact that the UAV and the target move deterministically to a surrounding sector; • Observation function, $O_i : S \times A \times \Omega_i \rightarrow [0,1]$ gives the likelihood that the UAV will be informed of the correct row in which the target is located; and • Reward function, $R_i : S \times A \rightarrow [0,1]$ formalizes the goal of spotting the reconnaissance target. If the UAV moves to a sector containing a target , we assume that the target is spotted and the game ends.

Because the target may move, it is beneficial for $i$ to anticipate $T$'s actions. Thus, the UAV tracks some possible beliefs that $T$ may have about the location of the UAV. We assume that the UAV is aware of $T$'s objectives that conflict with its own, the probabilities of its observations, and therefore $T$'s frame. We point out the size and complexity of this problem, involving 36 physical states, 5 actions and 3 observations for each agent.

Analogously to the previous problem sets, we measured the quality of the estimation provided by the I-PF for this larger problem. In Fig. 14($b$), we show the KL-Divergence of the approximate distribution as the number of particles allocated to the I-PF are increased. The KL-Divergence decreases rapidly as we increase the number of particles for both level 1 and 2 beliefs. However,

Doshi & Gmytrasiewicz

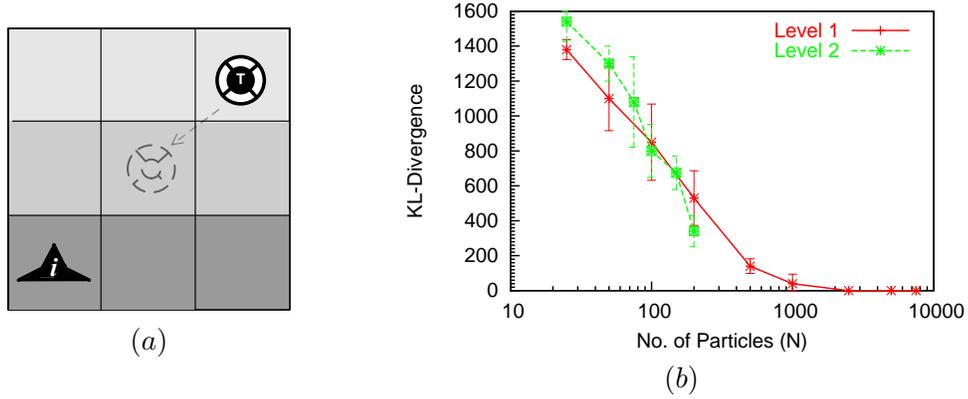

(a)

(b)

| Belief | Method | N=500 | N=1000 |
|--------|--------|-------|--------|
| Level 1 | I-PF | 2.929s ± 0.894s | 5.251s ± 0.492s |
| | Grid based | 3m 37.07s ± 4.22s | 7m 16.42s ± 0.27s |
| | | **N=100** | **N=200** |
| Level 2 | I-PF | 2m 55.52s ± 25.61s | 11m 10.43s ± 56.724s |

(c)

Figure 14: (a) The operating theater of the UAV $i$. The problem may be flexibly scaled by adding more targets and sectors. (b) The posterior obtained from the I-PF approaches the exact as the number of particles increase. (c) Comparison of the average running times of our numerical integration and I-PF implementations on the same platform (*Xeon, 3.0GHz, 2GB RAM, Linux*).

notice that the magnitude of the divergence is larger for lower numbers of particles in comparison to the previous problems. This is, in part, due to the larger state space of the problem and demonstrates that the I-PF does not fully address the curse of dimensionality. Thus, many more particles are needed to reach comparable levels of divergence. We also show a comparison of the run times of the I-POMDP belief update implemented using the I-PF and grid based numerical integration in the table in Fig. 14(c). An identical number of particles and grid points were selected, which provided comparable qualities of the estimations. We were unable to run the numerical integration implementation for level 2 beliefs for this problem.

## 9. Value Iteration on Sample Sets

Because the I-PF represents the belief of agent $i$, $b_{i,l}$, using a set of $N$ particles, $\widetilde{b}_{i,l}$, a value function backup operator which operates on samples is needed. Let $\widetilde{H}$ denote the required backup operator, and $\widetilde{U}$ the approximate value function, then the backup operation, $\widetilde{U}^t = \widetilde{H}\widetilde{U}^{t-1}$, is:

$$\widetilde{U}^t(\langle \widetilde{b}_{i,l}, \widehat{\theta}_i \rangle) = \max_{a_i \in A_i} \left\{ \frac{1}{N} \sum_{is^{(n)} \in \widetilde{b}_{i,l}} ER_i(is^{(n)}, a_i) + \gamma \sum_{o_i \in \Omega_i} Pr(o_i | a_i, \widetilde{b}_{i,l}) \widetilde{U}^{t-1}(\langle \text{I-PF}(\widetilde{b}_{i,l}, a_i, o_i), \widehat{\theta}_i \rangle) \right\}$$

(8)

where $ER_i(is^{(n)}, a_i) = \sum_{a_j} R_i(s^{(n)}, a_i, a_j) Pr(a_j | \theta_j^{(n)})$, $\gamma$ is the discount factor and I-PF$(\cdot)$ is the algorithm shown in Fig. 4. The set of optimal action(s) at a given approximate belief, OPT$(\langle \widetilde{b}_{i,l}, \widehat{\theta}_j \rangle)$, is then calculated by returning the action(s) that have the maximum value:

$$OPT(\langle \widetilde{b}_{i,l}, \widehat{\theta}_i \rangle) = \underset{a_i \in A_i}{argmax} \left\{ \begin{array}{l} \frac{1}{N} \sum_{is^{(n)} \in \widetilde{b}_{i,l}} ER_i(is^{(n)}, a_i) + \gamma \sum_{o_i \in \Omega_i} Pr(o_i | a_i, \widetilde{b}_{i,l}) \\ \times \widetilde{U}^t(\langle \text{I-PF}(\widetilde{b}_{i,l}, a_i, o_i), \widehat{\theta}_i \rangle) \end{array} \right\}$$

(9)

Equations 8 and 9 are analogous to Eqs. 4 and 5 respectively, with exact integration replaced by Monte Carlo integration, and the exact belief update replaced with the I-PF. Note that $\widetilde{H} \to H$ as $N \to \infty$.

The algorithm for computing an approximately optimal finite horizon policy tree given an initial belief using value iteration when $l > 0$ is shown in Fig. 18 in Appendix B.

## 9.1 Convergence and Error Bounds

The use of randomizing techniques such as PFs means that value iteration does not necessarily converge. This is because, unlike the exact belief update, posteriors generated by the PF with finitely many particles are not guaranteed to be identical for identical input. The non-determinism of the approximate belief update rules out isotonicity and contraction for $\widetilde{H}$ as $N \to \infty$. [6]

Our inability to guarantee convergence of value iteration implies that we must approximate an infinite horizon policy with the approximately optimal finite horizon policy. Let $U^*$ be the value of the optimal infinite horizon policy, $\widetilde{U}^t$ be the value of the approximate and $U^t$ be the value of the optimal $t$-horizon policy tree. Then the error bound (using the supremum norm $|| \cdot ||$) is, $||U^* - \widetilde{U}^t|| = ||U^* - U^t + U^t - \widetilde{U}^t|| \leq ||U^* - U^t|| + ||U^t - \widetilde{U}^t||$ (triangle inequality). Note that the first term, $||U^* - U^t||$, is bounded by $\gamma^t ||U^* - U^0||$. The bound for the second term is calculated below:

$$\begin{aligned} \mathcal{E}^t &= ||\widetilde{U}^t - U^t|| \\ &= ||\widetilde{H}\widetilde{U}^{t-1} - HU^{t-1}|| \\ &= ||\widetilde{H}\widetilde{U}^{t-1} - H\widetilde{U}^{t-1} + H\widetilde{U}^{t-1} - HU^{t-1}|| && \text{(add zero)} \\ &\leq ||\widetilde{H}\widetilde{U}^{t-1} - H\widetilde{U}^{t-1}|| + ||H\widetilde{U}^{t-1} - HU^{t-1}|| && \text{(triangle inequality)} \\ &\leq ||\widetilde{H}\widetilde{U}^{t-1} - H\widetilde{U}^{t-1}|| + \gamma ||\widetilde{U}^{t-1} - U^{t-1}|| && \text{(contracting } H) \\ &\leq ||\widetilde{H}\widetilde{U}^{t-1} - H\widetilde{U}^{t-1}|| + \gamma \mathcal{E}^{t-1} \end{aligned}$$

We turn our attention to calculating $||\widetilde{H}\widetilde{U}^{t-1} - H\widetilde{U}^{t-1}||$. In the analysis that follows we focus on level 1 beliefs. Let $\widehat{U}^t = H\widetilde{U}^{t-1}$, $\widetilde{U}^t = \widetilde{H}\widetilde{U}^{t-1}$, and $b_{i,1}$ be the singly nested belief where the worst error is made: $b_{i,1} = \underset{b_{i,1} \in B_{i,1}}{\operatorname{argmax}} |\dot{U}^t - \widetilde{U}^t|$. Let $\widetilde{\alpha}$ be the policy tree (alpha vector) that is optimal at $\widetilde{b}_{i,1}$ (the sampled estimate of $b_{i,1}$), and $\dot{\alpha}$ be the policy tree that is optimal at $b_{i,1}$. We will use Chernoff-Hoeffding (C-H) upper bounds (Theorem A.1.4, pg 265 in Alon & Spencer, 2000) [7], a well-known tool for analyzing randomized algorithms, to derive a confidence threshold $1 - \delta$ at which the observed estimate, $\widetilde{U}^t_{\widetilde{\alpha}}$, is within $2\epsilon$ of the true estimate $\dot{U}^t_{\dot{\alpha}}$ ($= E[\dot{\alpha}]$):

$$Pr(\widetilde{U}^t_{\widetilde{\alpha}} > \dot{U}^t_{\dot{\alpha}} + \epsilon) \leq e^{-2N\epsilon^2/(\widetilde{\alpha}_{max} - \widetilde{\alpha}_{min})^2}$$
$$Pr(\widetilde{U}^t_{\widetilde{\alpha}} < \dot{U}^t_{\dot{\alpha}} - \epsilon) \leq e^{-2N\epsilon^2/(\widetilde{\alpha}_{max} - \widetilde{\alpha}_{min})^2}$$

---

6. One may turn PFs into deterministic belief update operators (de-randomization) by generating several posteriors from the same input. A representative posterior is then formed by taking a convex combination of the different posteriors. For example, Thrun (2000) uses a $k$-nearest neighborhood approach for this purpose.

We may write,

$$Pr(\widetilde{U}^t_{\widetilde{\alpha}} > \dot{U}^t_{\dot{\alpha}} + \epsilon \text{ OR } \widetilde{U}^t_{\widetilde{\alpha}} < \dot{U}^t_{\dot{\alpha}} - \epsilon) = Pr(\widetilde{U}^t_{\widetilde{\alpha}} > \dot{U}^t_{\dot{\alpha}} + \epsilon) + Pr(\widetilde{U}^t_{\widetilde{\alpha}} < \dot{U}^t_{\dot{\alpha}} - \epsilon)$$
$$- Pr(\widetilde{U}^t_{\widetilde{\alpha}} > \dot{U}^t_{\dot{\alpha}} + \epsilon \text{ AND } \widetilde{U}^t_{\widetilde{\alpha}} < \dot{U}^t_{\dot{\alpha}} - \epsilon)$$

As the last term is zero, the equation becomes:

$$Pr(\widetilde{U}^t_{\widetilde{\alpha}} > \dot{U}^t_{\dot{\alpha}} + \epsilon \text{ OR } \widetilde{U}^t_{\widetilde{\alpha}} < \dot{U}^t_{\dot{\alpha}} - \epsilon) \leq 2e^{-2N\epsilon^2/(\widetilde{\alpha}_{max} - \widetilde{\alpha}_{min})^2}$$

We may replace $Pr(\widetilde{U}^t_{\widetilde{\alpha}} > \dot{U}^t_{\dot{\alpha}} + \epsilon \text{ OR } \widetilde{U}^t_{\widetilde{\alpha}} < \dot{U}^t_{\dot{\alpha}} - \epsilon)$ with $1 - Pr(\dot{U}^t_{\dot{\alpha}} - \epsilon \leq \widetilde{U}^t_{\widetilde{\alpha}} \leq \dot{U}^t_{\dot{\alpha}} + \epsilon)$. After some simple operations, the above inequality becomes:

$$Pr(\dot{U}^t_{\dot{\alpha}} - \epsilon \leq \widetilde{U}^t_{\widetilde{\alpha}} \leq \dot{U}^t_{\dot{\alpha}} + \epsilon) \geq 1 - 2e^{-2N\epsilon^2/(\widetilde{\alpha}_{max} - \widetilde{\alpha}_{min})^2}$$

Let the probability that $\widetilde{U}^t_{\widetilde{\alpha}}$ is within $2\epsilon$ of the true estimate $\dot{U}^t_{\dot{\alpha}}$, be at least $1 - \delta$. Then we have:

$$1 - \delta = 1 - 2e^{-2N\epsilon^2/(\widetilde{\alpha}_{max} - \widetilde{\alpha}_{min})^2}$$

With a confidence probability of at least $1 - \delta$, the error bound is:

$$\epsilon = \sqrt{\frac{(\widetilde{\alpha}_{max} - \widetilde{\alpha}_{min})^2 ln(2/\delta)}{2N}} \qquad (10)$$

where $\widetilde{\alpha}_{max} - \widetilde{\alpha}_{min}$ may be loosely upper bounded as $\frac{R_{max} - R_{min}}{1 - \gamma}$. Note that Eq. 10 can also be used to derive the number of particles, $N$, for some given $\delta$ and $\epsilon$. To get the desired bound, we note that with at least probability $1 - \delta$ our error bound is $2\epsilon$ and with probability at most $\delta$ the worst possible suboptimal behavior may result: $||\widetilde{H}\widetilde{U}^{t-1} - H\widetilde{U}^{t-1}|| \leq (1 - \delta)2\epsilon + \delta\frac{R_{max} - R_{min}}{1 - \gamma}$. The final error bound now obtains:

$$\begin{aligned} \mathcal{E}^t &\leq (1 - \delta)2\epsilon + \delta\frac{R_{max} - R_{min}}{1 - \gamma} + \gamma\mathcal{E}^{t-1} \\ &= (1 - \delta)\frac{2\epsilon(1 - \gamma^t)}{1 - \gamma} + \delta\frac{(R_{max} - R_{min})(1 - \gamma^t)}{(1 - \gamma)^2} \quad \text{(geometric series)} \end{aligned} \qquad (11)$$

where $\epsilon$ is as defined in Eq. 10.

---

7. At horizon $t$, samples in $\widetilde{b}_{i,1}$ are i.i.d. However, at horizons less than $t$, the samples are generated by the I-PF and exhibit limited statistical independence, but independent research (Schmidt, Siegel, & Srinivasan, 1995) reveals that C-H bounds still apply.

**Proposition 3** (Error Bound). *For a singly nested $t$-horizon I-POMDP$_{i,1}$, the error introduced by our approximation technique is upper bounded and is given by:*

$$||\widetilde{U}^t - U^t|| \leq (1-\delta)\frac{2\epsilon(1-\gamma^t)}{1-\gamma} + \delta\frac{(R_{max} - R_{min})(1-\gamma^t)}{(1-\gamma)^2}$$

*where $\epsilon$ is as defined in Eq. 10.*

At levels of belief nesting greater than one, $j$'s beliefs are also approximately represented using samples. Hence the approximation error is not only due to the sampling, but also due to the possible incorrect prediction of $j$'s actions based on its approximate beliefs. Since even a slight deviation from the exact belief may lead to an action that turns out to be the worst in value when compared to the optimal action, it seems difficult to derive bounds that are useful – tighter than the usual difference between the best and worst possible behavior ($\frac{R_{max} - R_{min}}{(1-\gamma)^2}$) – for this case.

### 9.2 Computational Savings

Since the complexity of solving I-POMDPs is dominated by the complexity of solving the models of other agents we analyze the reduction in the number of agent models that must be solved. In a $K+1$-agent setting with the number of particles bounded by $N$ at each level, each particle in $\widetilde{b}_{k,l}^{t-1}$ of level $l$ contains $K$ models all of level $l-1$. Solution of each of these level $l-1$ models requires solution of the lower level models recursively. The upper bound on the number of models that are solved is $O((KN)^{l-1})$. Given that there are $K$ level $l-1$ models in a particle, and $N$ such possibly distinct particles, we need to solve $O((KN)^l)$ models. Our upper bound on the number of models to be solved is polynomial in $K$ for a fixed nesting level. This can be contrasted with $O((K|\Theta_*|^K)^l)$ models that need to be solved in the exact case, which is exponential in $K$. Here, among the spaces of models of all agents, $\Theta_*$ is the largest space and is theoretically countably infinite. Typically, $N \ll |\Theta_*|^K$, resulting in a substantial reduction in computation. However, note that the total number of particles is exponential in the nesting level, $l$. This makes solutions for large nesting levels still intractable.

## 10. Empirical Performance

The goal of our experimental analysis is to demonstrate empirically, (a) the reduction in error with increasing sample complexity, and (b) savings in computation time when the approximation technique is used. We again use the multiagent tiger problem introduced previously, and a multiagent version of the machine maintenance (MM) problem (see Appendix A) as test problems. While the single-agent versions of these problems are simple, their multiagent versions are sufficiently complex so as to motivate the use of approximation techniques to solve them. Additionally, we demonstrate that our approach scales to larger problems by applying it to the UAV reconnaissance problem as well.

### 10.1 Multiagent Tiger and Machine Maintenance Problems

To demonstrate the reduction in error, we construct performance profiles showing an increase in performance as more computational resources – in this case particles – are allocated to the approximation algorithm. In Figs. 15(a) and (c) we show the performance profile curves when agent $i$'s prior belief is the level 1 belief described previously in Fig. 3(a), and suitably modified for the MM

**Multiagent Tiger Problem**

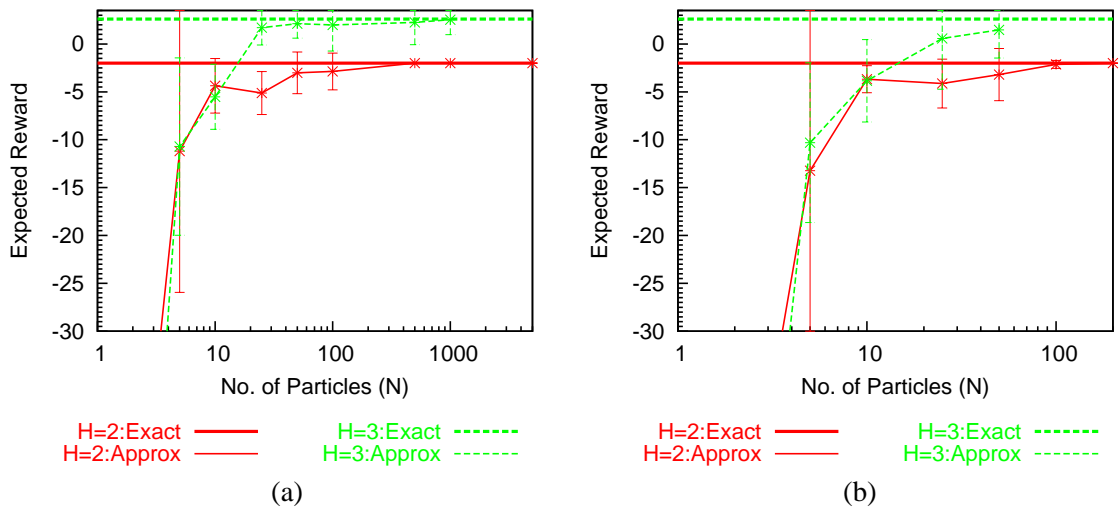

**Multiagent Machine Maintenance Problem**

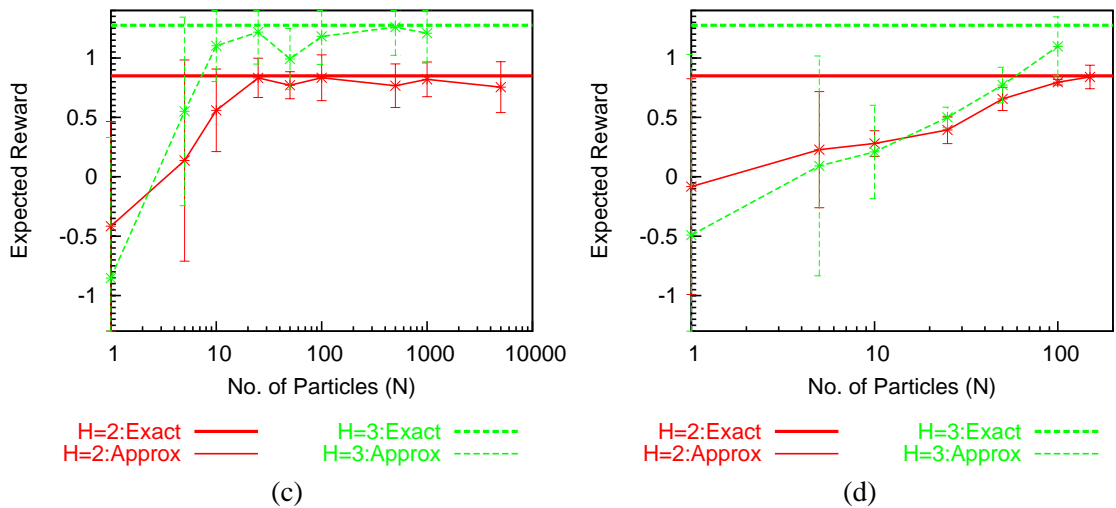

Figure 15: Anytime performance profiles: The multiagent tiger problem using the (a) level 1, and (b) level 2 belief as the prior for agent $i$. The multiagent MM using the (c) level 1, and (d) level 2 belief as $i$'s prior. The approximate policies gradually improve as we employ an increasing number of particles.

problem. As expected the average rewards for both, horizon 2 and 3 approach the optimal expected reward as the number of particles increases. We show the analogous plots for a level 2 belief in Figs. 15(b) and (d). In each of these cases the average of the rewards accumulated by $i$ over a 2 and 3 horizon policy tree (computed using the APPROXPOLICY algorithm in Fig. 18) while playing against agent $j$ in simulated tiger and MM problems were plotted. To compensate for the randomness in sampling, we generated $i$'s policy tree 10 times independently of each other, and averaged over 100 runs each time. Within each run, the location of the tiger and $j$'s prior beliefs were sampled according to $i$'s prior belief. $j$'s policy was then computed using the algorithm in Fig. 18.

| Problem | Error | $t = 2$ | | $t = 3$ | |
|---|---|---|---|---|---|
| | | **N=100** | **N=1000** | **N=100** | **N=1000** |
| Multiagent tiger | Obs. | 5.61 | 0 | 4.39 | 2.76 |
| | $\mathcal{E}^t$ | 64.73 | 41.53 | 120.90 | 77.57 |
| | Worst | 209.00 | | 298.10 | |
| Multiagent MM | Obs. | 0.28 | 0.23 | 0.46 | 0.40 |
| | $\mathcal{E}^t$ | 4.58 | 2.05 | 8.79 | 3.64 |
| | Worst | 8.84 | | 12.61 | |

Table 3: Comparison of the worst case observed errors, our theoretical error bounds and the trivial error bound ($\frac{R_{max} - R_{min}}{(1-\gamma)^2}$).

In Table 3, we compare the empirically determined error bound – difference between the optimal expected reward and the worst observed expected reward – with the theoretical error bound ($\delta$=0.1, $\gamma$=0.9) from Section 9.1, for horizons 2 and 3. The theoretical error bounds appear loose due to the worst-case nature of our analysis but (expectedly) are much tighter than the trivial worst bounds, and become better as the number of particles increases.

| Problem | Method | Run times | | | |
|---|---|---|---|---|---|
| | | $t = 2$ | $t = 3$ | $t = 4$ | $t = 5$ |
| Multiagent tiger | Grid | 37.84s $\pm$ 0.6s | 11m 22.25s $\pm$ 1.34s | * | * |
| | SB | 1.44s $\pm$ 0.05s | 1m 44.29s $\pm$ 0.6s | 19m 16.88s $\pm$ 17.5s | * |
| Multiagent MM | Grid | 5m 26.57s $\pm$ 0.07s | 20m 45.69s $\pm$ 0.29s | * | * |
| | SB | 5.75s $\pm$ 0.01s | 34.52s $\pm$ 0.01s | 3m 24.9s $\pm$ 0.04s | 17m 58.39s $\pm$ 0.57s |

Table 4: Run times on a Pentium IV 2.0 GHz, 2.0GB RAM and Linux. * = program ran out of memory.

Table 4 compares the average run times of our sample-based approach (SB) with the grid based approach, for computing policy trees of different horizons starting from the level 1 belief. The values of the policy trees generated by the two approaches were similar. The run times demonstrate the impact of the curse of dimensionality on the grid based method as shown by the higher run times for the MM problem in comparison to the tiger problem. Our I-PF based implementation though not immune to this curse reduces its impact, but is affected by the curse of history, as illustrated by the higher run times for the tiger problem (branching factor of the reachability tree: $|A_i||\Omega_i| = 18$) compared to the MM problem (branching factor: $|A_i||\Omega_i| = 8$). We were unable to compute the solutions using the grid based implementation for both the problems for horizons beyond 3.

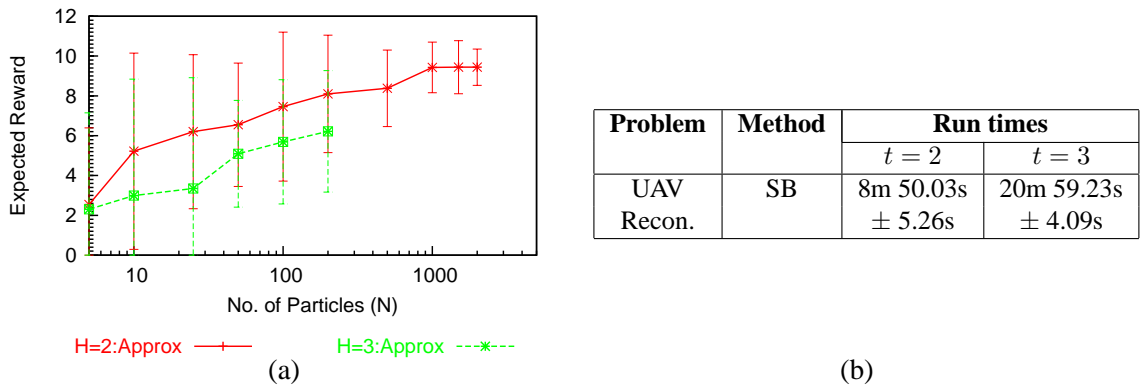

Figure 16: (a) Anytime performance profile for the UAV reconnaissance problem for horizons 2 and 3. Notice that the profile flattens when the number of particles reaches 1,000 and greater, thereby indicating that the corresponding average reward is close to optimal. (b) Run times for obtaining a policy on a Xeon 3.0 GHz, 2.0GB RAM and Linux. For horizon 2, we used 500 particles while 100 particles were used for horizon 3. As we see in (a), rewards of policies at these numbers of particles are not much less than the converged values.

## 10.2 UAV Reconnaissance Problem

We evaluate the performance of our approach on the UAV reconnaissance problem, which we introduced previously in Section 8.3. As we mentioned, this is a larger problem consisting of 36 physical states, 5 actions and 3 observations. We show the level 1 performance profile for this problem in Fig. 16(a) for horizons 2 and 3. Due to the size of the problem, we were unable to compute the exact value of the optimal policies. As before, each data point is the average reward obtained by simulating a horizon 2 or 3 policy in a two-agent setting. Agent $i$'s initial belief is one according to which it is uncertain about the physical state and models of the other agent. We observe that the profiles tend to flatten as the number of particles increases. The corresponding expected reward is therefore close to the optimal value of the policies.

We also show the time taken to generate a good quality horizon 2 and 3 policy on average. They indicate that while it is indeed possible to obtain (approximate) policies for large problems, the times needed are somewhat large. Notice that these times are significantly greater than the run times for the multiagent tiger and MM problems. This is, in part, due to the larger state space and, as we show later in this article, in part due to the larger numbers of actions and observations of each agent.

## 11. Sampling the Look Ahead Reachability Tree

Although we were able to solve I-POMDPs for large state spaces, we were unable to generate solutions for large horizons. The main reason for this is the exponential growth of the look ahead reachability tree with increasing horizons; we referred to this as the curse of history. At some time step $t$, there could be $(|A_i||\Omega_i|)^{t-1}$ reachable belief states of the agent $i$. For example, in the multiagent tiger problem, at the second time step there could be 18 possible belief states, 324 of them at the third time step, and more than 0.1 million at the fifth time step.

To mitigate the curse of history, we reduce the branching factor of the look ahead reachability tree by sampling from the possible observations that the agent may receive. While this approach does not completely address the curse of history, it beats back the impact of this curse substantially. During the reachability tree expansion phase, the agent's actions are used to propagate its belief. Observations are then sampled at each propagated belief, $o_i^t \sim Pr(\Omega_i | a_i^{t-1}, \tilde{b}_{i,l}^{t-1})$ where $\tilde{b}_{i,l}^{t-1}$ is the propagated belief. During the value iteration phase, value is backed up the (possibly) partial reachability tree, and the agent performs the action that is optimal at the root node of the tree. For convenience, let us label this approach as *reachability tree sampling* (RTS).

RTS shares its conceptual underpinnings with the belief expansion models of PBVI (Pineau et al., 2006), but differs in that our method is applicable to online policy tree generation for I-POMDPs, compared to PBVI's use in offline policy generation for POMDPs. It is also similar to the sparse sampling technique proposed for selecting actions for exploration during reinforcement learning (Kearns et al., 2002) and for online planning in POMDPs (Ross et al., 2008) by sampling the look ahead trees. A distinction is that we focus on sampling observations given the propagated multiagent beliefs, while the latter approaches focused on settings of single agent learning. Consequently, the process of computing the sampling distribution and the experimental settings differ.

## 11.1 Computational Savings

We consider the computational savings that result from sampling observations in the look ahead reachability tree. If we are sampling $N_{\Omega_i} < |\Omega_i|$ observations at each propagated belief within the reachability tree, then at some time step $t$, we will obtain $(|A_i||N_{\Omega_i}|)^{t-1}$ possible belief states, assuming the worst case occurs and we end up sampling $N_{\Omega_i}$ distinct observations. This is in comparison to the $(|A_i||\Omega_i|)^{t-1}$ belief states in the complete reachability tree. Typically, as our experiments demonstrate, the number of distinct sampled observations is less than $|\Omega_i|$, resulting in significant computational savings.

| Method | Run times | | | | | |
|---|---|---|---|---|---|---|
| | $t = 2$ | $t = 3$ | $t = 4$ | $t = 5$ | $t = 6$ | $t = 7$ |
| SB-No-RTS | 1.44s $\pm$ 0.05s | 1m 44.29s $\pm$ 0.6s | 19m 16.88s $\pm$ 17.5s | * | * | * |
| SB-RTS | 0.86s $\pm$ 0.02s | 15.17s $\pm$ 1.6s | 2m 52.9s $\pm$ 6.51s | 4m 13.43s $\pm$ 27.51s | 7m 29.9s $\pm$ 47.98s | 11m 51.57s $\pm$ 20.05s |

Table 5: Run times for the multiagent tiger problem on a Pentium IV 2.0 GHz, 2.0GB RAM and Linux. * = program ran out of memory.

As an illustration of the computational savings we compare the run times of computing the policy tree for the multiagent tiger (Table 5) and the UAV reconnaissance (Table 6) problems for singly-nested beliefs. We compare value iteration in which the reachability tree is sampled (SB-RTS) with value iteration that does no reachability tree sampling (SB-No-RTS), the algorithm for which is given in Fig. 18. For SB-RTS in the multiagent tiger problem, we sampled eight times from the observation distribution up to the fifth horizon and six times thereafter. For both the algorithms, we used a similar number of particles in the I-PF. As the tiger problem has a total of 6 observations, not only does the SB-RTS compute the policy faster, we were able to compute it up to *seven* time horizons. When compared with the performance of SB-No-RTS, these results demonstrate that the approach of sampling the reachability tree could yield significant computational savings.

| Method | Run times | | |
|--------|-----------|--------|--------|
| | $t = 2$ | $t = 3$ | $t = 4$ |
| SB-No-RTS | 8m 50.03s $\pm$ 5.26s | 20m 59.23s $\pm$ 4.09s | * |
| SB-RTS | 8m 23.86s $\pm$ 20.46s | 19m 25.54s $\pm$ 89.08s | 121m 16.2s $\pm$ 592s |

Table 6: Run times for the UAV reconnaissance problem on a Xeon 3.0 GHz, 2.0GB RAM and Linux. * = program ran out of memory.

However, as we see in Table 6, the approach does not yield significant savings in the context of the UAV problem for small horizons. Because the space of observations in the problem is small (3 distinct observations), a majority of these are often selected on sampling. Hence, for smaller horizons such as 2 and 3, we did not observe a significant decrease in the size of the look ahead reachability tree. However, the reduction in the look ahead trees for horizon 4 is enough to allow the computation of the corresponding policy, though the run time is considerable. We were unable to obtain it for the case with no RTS. Thus, the UAV problem reveals an important limitation of this technique – it may not provide significant computational savings when the space of observations is small, particularly for small horizons.

## 11.2 Empirical Performance

We present the performance profiles in Fig. 17 for the multiagent tiger problem when partial look ahead reachability trees are built by sampling the observations. Similar to our previous experiments, the performance profiles reflect the average of the rewards accumulated by following the action prescribed by the root of the approximate policy tree that is built online. We plot the average reward accumulated by $i$ over 10 independent trials consisting of 100 runs each, as the number of the observation samples, $N_{\Omega_i}$ are gradually increased. Within each run, the location of the tiger and $j$'s prior beliefs were sampled according to $i$'s prior level 1 belief. Since we have combined RTS with the I-PF, in addition to varying $N_{\Omega_i}$, we also vary the number of particles, $N_p$, employed to approximate the beliefs. As expected of performance profiles, the expected reward initially increases sharply, before flattening out as $N_{\Omega_i}$ becomes large and the sampled observation distribution reaches the true one. Reflecting intuition, the plots for $N_p = 100$ exhibit slightly better expected rewards on average than those for $N_p = 50$. While the increase is not large, we note that it is consistent across all observation samples. We also obtained the average reward over a similar number of trials when a random policy (null hypothesis) is used for $i$. For horizon 3, the random policy gathered an average reward of -84.785 ($\pm$ 37.9847), and -108.5 ($\pm$ 41.56) for horizon 4. Even for a small number of observation samples, RTS does significantly better than the random policy thereby demonstrating the usefulness of partial tree expansion. However, we note that a random policy is a poor baseline for comparison and is used due to an absence of other similar approximations for I-POMDPs.

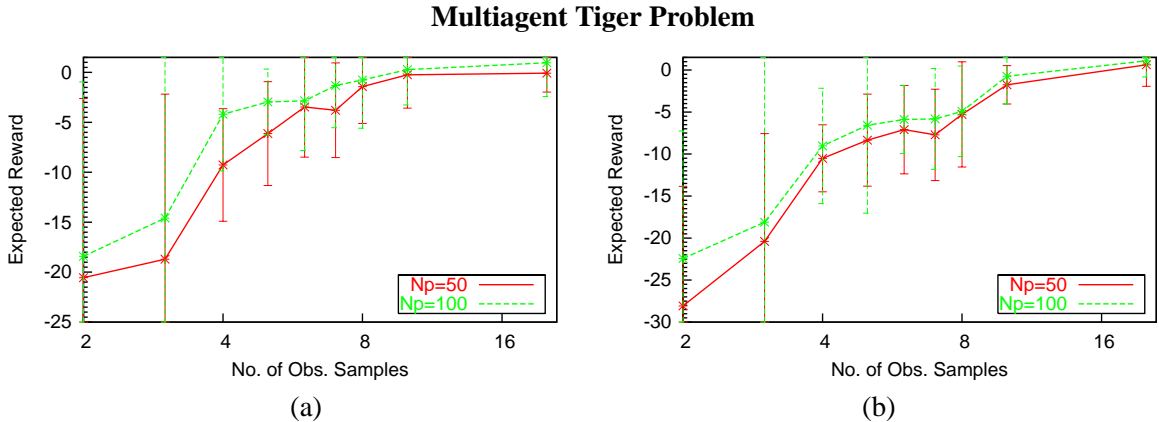

Figure 17: Performance profiles for the multiagent tiger problem for (a) horizon 3, and for (b) horizon 4 when the look ahead tree is built by sampling observations.

Due to the small number of observations in the MM and UAV reconnaissance problems, we did not observe significant increases in the expected reward as more observations are sampled. Hence, we do not show the performance profiles for these problems.

We observed that the empirical expected reward is close to the optimal expected reward when only a few distinct observations were sampled while building the reachability tree. This observation when combined with the computational savings demonstrated in Section 11.1 indicate that our approximation approach is viable. Additionally, by varying the parameters $N_p$ and $N_{\Omega_i}$, we can flexibly, though only partially, control the effects of both the curses of dimensionality and history, respectively, on the solutions according to the application requirements. An interesting line of future work is to investigate the interplay of these parameters.

## 12. Discussion

We described randomized methods for obtaining approximate solutions to finitely nested I-POMDPs based on a novel generalization of particle filtering to multiagent settings. The generalization is not straightforward because we are confronted with an interactive belief hierarchy in multiagent settings. We proposed the interactive particle filter which descends the levels of interactive belief hierarchies, and samples and propagates beliefs at each level. While sampling methods such as particle filters are unable to completely avoid the curse of dimensionality, they serve to focus the computational resources on those elements of the model that matter the most to the agent.

However, the interactive particle filter does not address the policy space complexity. Though value iteration using sample sets is not guaranteed to converge asymptotically, we established useful error bounds for singly-nested I-POMDPs. Their generalization to multiply-nested beliefs has proved to be difficult but we continue to investigate it. We provided performance profiles for the multiagent tiger and the machine maintenance problems, and demonstrated scalability using the larger UAV reconnaissance problem.

The experiments show that the approach saves on computation over the space of models but it does not scale (usefully) to large values of time horizons and needs to be combined with methods that deal with the curse of history. In order to reduce the impact of the curse of history, we proposed to sample observations while constructing the look ahead tree during the reachability analysis phase of policy computation. This sparse sampling technique effectively reduces the branching factor of the tree and allows computation of solutions for larger horizons as we demonstrated.

A method to further scale the approximation technique is to pick a subset of actions in addition to sampling the observations while building the reachability tree. This further dampens the exponential growth of the reachability tree with increasing horizons, and permits solutions for larger horizons. However, this approach must be used cautiously – we do not want to leave out critical actions from the policy.

Specific approaches to speeding up the computation also remain to be explored. As we mentioned, the number of particles in the interactive particle filter grows exponentially with the number of nesting levels. In this regard, can we assign monotonically decreasing number of particles to represent beliefs nested at deeper levels exploiting an insight from cognitive psychology that beliefs nested at deeper levels are less likely to influence the optimal policy? Thus, if we decrease the particles sampled at the rate of $r < 1$, there will be no more than $N_p/(1-r)$ particles in total, resulting in computational savings.

Finally, in regards to the appropriate strategy level to use for nested models, we note the analogy with classical POMDPs, and the amount of detail and modeling information included therein. Adding more nested level of modeling is analogous to including more details in the POMDP formulation. Then, the solution to an I-POMDP is optimal given the level of detail included in the model, just like for classical POMDPs.

## Acknowledgments

This research is supported in part by grant #FA9550-08-1-0429 from AFOSR and in part by grants IRI-9702132 and IRI-0119270 from NSF. Versions of parts of this article have previously appeared in (Doshi & Gmytrasiewicz, 2005a) and in (Doshi & Gmytrasiewicz, 2005b). We acknowledge all the reviewers for their useful comments.

## Appendix A. Multiagent Machine Maintenance Problem

We extend the traditional single agent version of the machine maintenance (MM) problem (Smallwood & Sondik, 1973) to a two-agent cooperative version. The original MM problem involved a machine containing two internal components operated by a single agent. Either one or both components of the machine may fail spontaneously after each production cycle. If an internal component has failed, then there is some chance that when operating upon the product, it will cause the product to be defective. The agent may choose to manufacture the product (M) without examining it, examine the product (E), inspect the machine (I), or repair it (R) before the next production cycle. On an examination of the product, the subject may find it to be defective. Of course, if more components have failed, then the probability that the product is defective is greater.

The transition function, observation functions, and the reward functions for the two agents, $i$ and $j$, are as shown in Table 7. Apart from including two agents that operate on the machine during the production cycle, we increased the nondeterminism of the original problem to make it more realistic. This also has the beneficial effect of producing a richer policy structure.

| $\langle a_i, a_j \rangle$ | State | 0-fail | 1-fail | 2-fail |
|---|---|---|---|---|
| $\langle M/E,M/E \rangle$ | 0-fail | 0.81 | 0.18 | 0.01 |
| $\langle M/E,M/E \rangle$ | 1-fail | 0.0 | 0.9 | 0.1 |
| $\langle M/E,M/E \rangle$ | 2-fail | 0.0 | 0.0 | 1.0 |
| $\langle M,I/R \rangle$ | 0-fail | 1.0 | 0.0 | 0.0 |
| $\langle M,I/R \rangle$ | 1-fail | 0.95 | 0.05 | 0.0 |
| $\langle M,I/R \rangle$ | 2-fail | 0.95 | 0.0 | 0.05 |
| $\langle E,I/R \rangle$ | 0-fail | 1.0 | 0.0 | 0.0 |
| $\langle E,I/R \rangle$ | 1-fail | 0.95 | 0.05 | 0.0 |
| $\langle E,I/R \rangle$ | 2-fail | 0.95 | 0.0 | 0.05 |
| $\langle I/R,* \rangle$ | 0-fail | 1.0 | 0.0 | 0.0 |
| $\langle I/R,* \rangle$ | 1-fail | 0.95 | 0.05 | 0.0 |
| $\langle I/R,* \rangle$ | 2-fail | 0.95 | 0.0 | 0.05 |

Transition function for agents $i$ and $j$: $T_i = T_j$

| $\langle a_i, a_j \rangle$ | State | not-defective | defective |
|---|---|---|---|
| $\langle M,M/E \rangle$ | * | 0.5 | 0.5 |
| $\langle M,I/R \rangle$ | * | 0.95 | 0.05 |
| $\langle E,M/E \rangle$ | 0-fail | 0.75 | 0.25 |
| $\langle E,M/E \rangle$ | 1-fail | 0.5 | 0.5 |
| $\langle E,M/E \rangle$ | 2-fail | 0.25 | 0.75 |
| $\langle E,I/R \rangle$ | * | 0.95 | 0.05 |
| $\langle I/R,* \rangle$ | * | 0.95 | 0.05 |

| $\langle a_i, a_j \rangle$ | State | not-defective | defective |
|---|---|---|---|
| $\langle M/E,M \rangle$ | * | 0.5 | 0.5 |
| $\langle I/R,M \rangle$ | * | 0.95 | 0.05 |
| $\langle M/E,E \rangle$ | 0-fail | 0.75 | 0.25 |
| $\langle M/E,E \rangle$ | 1-fail | 0.5 | 0.5 |
| $\langle M/E,E \rangle$ | 2-fail | 0.25 | 0.75 |
| $\langle I/R,E \rangle$ | * | 0.95 | 0.05 |
| $\langle *,I/R \rangle$ | * | 0.95 | 0.05 |

Observation functions for agents $i$ and $j$.

| $\langle a_i, a_j \rangle$ | 0-fail | 1-fail | 2-fail |
|---|---|---|---|
| $\langle M,M \rangle$ | 1.805 | 0.95 | 0.5 |
| $\langle M,E \rangle$ | 1.555 | 0.7 | 0.25 |
| $\langle M,I \rangle$ | 0.4025 | -1.025 | -2.25 |
| $\langle M,R \rangle$ | -1.0975 | -1.525 | -1.75 |
| $\langle E,M \rangle$ | 1.5555 | 0.7 | 0.25 |
| $\langle E,E \rangle$ | 1.305 | 0.45 | 0.0 |
| $\langle E,I \rangle$ | 0.1525 | -1.275 | -2.5 |
| $\langle E,R \rangle$ | -1.3475 | -1.775 | -2.0 |
| $\langle I,M \rangle$ | 0.4025 | -1.025 | -2.25 |
| $\langle I,E \rangle$ | 0.1525 | -1.275 | -2.5 |
| $\langle I,I \rangle$ | -1.0 | -3.00 | -5.00 |
| $\langle I,R \rangle$ | -2.5 | -3.5 | -4.5 |
| $\langle R,M \rangle$ | -1.0975 | -1.525 | -1.75 |
| $\langle R,E \rangle$ | -1.3475 | -1.775 | -2.0 |
| $\langle R,I \rangle$ | -2.5 | -3.5 | -4.5 |
| $\langle R,R \rangle$ | -4 | -4 | -4 |

| $\langle a_i, a_j \rangle$ | 0-fail | 1-fail | 2-fail |
|---|---|---|---|
| $\langle M,M \rangle$ | 1.805 | 0.95 | 0.5 |
| $\langle M,E \rangle$ | 1.555 | 0.7 | 0.25 |
| $\langle M,I \rangle$ | 0.4025 | -1.025 | -2.25 |
| $\langle M,R \rangle$ | -1.0975 | -1.525 | -1.75 |
| $\langle E,M \rangle$ | 1.555 | 0.7 | 0.25 |
| $\langle E,E \rangle$ | 1.305 | 0.45 | 0.0 |
| $\langle E,I \rangle$ | 0.1525 | -1.275 | -2.5 |
| $\langle E,R \rangle$ | -1.3475 | -1.775 | -2.0 |
| $\langle I,M \rangle$ | 0.4025 | -1.025 | -2.25 |
| $\langle I,E \rangle$ | 0.1525 | -1.275 | -2.5 |
| $\langle I,I \rangle$ | -1.0 | -3.00 | -5.00 |
| $\langle I,R \rangle$ | -2.5 | -3.5 | -4.5 |
| $\langle R,M \rangle$ | -1.0975 | -1.525 | -1.75 |
| $\langle R,E \rangle$ | -1.3475 | -1.775 | -2.0 |
| $\langle R,I \rangle$ | -2.5 | -3.5 | -4.5 |
| $\langle R,R \rangle$ | -4 | -4 | -4 |

Reward functions for agents $i$ and $j$.

Table 7: Transition, observation and reward functions for the multiagent MM problem.

## Appendix B. Algorithm for Value Iteration on Sample Sets

We show the algorithm for computing an approximately optimal finite horizon policy tree given an initial belief using value iteration when $l > 0$. When $l = 0$, the algorithm reduces to the POMDP policy tree computation which is carried out exactly.[8] The algorithm consists of the usual two steps: compute the look ahead reachability tree for horizon $T$ as part of the reachability analysis (see Section 17.5 of Russell & Norvig, 2003) in lines 2-6 and perform value backup on the reachability tree, in lines 7-28. The value of the beliefs at the leaves of the reachability tree is simply the one-step expected reward resulting from the best action.

---

8. For large problems, exact POMDP solutions may be replaced with approximate ones. But in doing so, our error bounds will no longer be applicable and those of the approximation technique will have to be considered.

---

**Function** APPROXPOLICY($\theta_k, l > 0$) **returns** $\Delta(A_k)$

1. $\widetilde{b}_{k,l}^0 \leftarrow \{is_k^{(n)}, n = 1 \ldots N | is_k^{(n)} \sim b_{k,l} \in \theta_k\}$       *//Initial sampled belief*

    <u>*Reachability Analysis*</u>

2. reach$(0) \leftarrow \widetilde{b}_{k,l}^0$
3. **for** $t \leftarrow 1$ **to** $T - 1$ **do**
4.      reach$(t) \leftarrow \phi$
5.      **for all** $\widetilde{b}_{k,l}^{t-1} \in$ reach$(t-1)$, $a_k \in A_k$, $o_k \in \Omega_k$ **do**
6.          reach$(t) \overset{\cup}{\leftarrow}$ I-PARTICLEFILTER$(\widetilde{b}_{k,l}^{t-1}, a_k, o_k, l)$

    *Dynamic Programming*

7. **for** $t \leftarrow T - 1$ **downto** $0$ **do**
8.      **for all** $\widetilde{b}_{k,l}^t \in$ reach$(t)$ **do**
9.          $\widetilde{U}^{T-t}(\langle \widetilde{b}_{k,l}^t, \widehat{\theta}_k \rangle) \leftarrow -\infty$, OPT$(\langle \widetilde{b}_{k,l}^t, \widehat{\theta}_k \rangle) \leftarrow \phi$
10.          **for all** $a_k \in A_k$ **do**
11.              $\widetilde{U}_{a_k}^{T-t}(\langle \widetilde{b}_{k,l}^t, \widehat{\theta}_k \rangle) \leftarrow 0$
12.              **for all** $is_k^{(n),t} = \langle s^{(n),t}, \theta_{-k}^{(n)} \rangle \in \widetilde{b}_{k,l}^t$ **do**
13.                  $Pr(A_{-k} | \theta_{-k}^{(n)}) \leftarrow$ APPROXPOLICY$(\theta_{-k}^{(n)}, l-1)$
14.              **for all** $a_{-k} \in A_{-k}$ **do**
15.                  $\widetilde{U}_{a_k}^{T-t}(\langle \widetilde{b}_{k,l}^t, \widehat{\theta}_k \rangle) \overset{+}{\leftarrow} \frac{1}{N} R(s^{(n),t}, a_k, a_{-k}) Pr(a_{-k} | \theta_{-k}^{(n)})$
16.              **if** $(t < T)$ **then**
17.                  **for all** $o_k \in \Omega_k$ **do**
18.                      sum $\leftarrow 0$, $\widetilde{b}_{k,l}^{t+1} \leftarrow$ reach$(t+1)[|\Omega_k | a_k + o_k]$
19.                      **for all** $is_k^{(n),t} = \langle s^{(n),t}, \theta_{-k}^{(n)} \rangle \in \widetilde{b}_{k,l}^t$ **do**
20.                          $Pr(A_{-k} | \theta_{-k}^{(n)}) \leftarrow$ APPROXPOLICY$(\theta_{-k}^{(n)}, l-1)$
21.                      **for all** $a_{-k} \in A_{-k}$, $s^{t+1} \in S_k$ **do**
22.                          sum $\overset{+}{\leftarrow} O_k(o_k | s^{t+1}, a_k, a_{-k}) Pr(is^{(n),t+1} | is^{(n),t}, a_k, a_{-k}) Pr(a_{-k} | \theta_{-k}^{(n)})$
23.                    $\widetilde{U}_{a_k}^{T-t}(\langle \widetilde{b}_{k,l}^t, \widehat{\theta}_k \rangle) \overset{+}{\leftarrow} \gamma \times \frac{1}{N} \times$ sum $\times \widetilde{U}^{T-t-1}(\widetilde{b}_{k,l}^{t+1})$
24.              **if** new value $\widetilde{U}_{a_k}^{T-t}(\langle \widetilde{b}_{k,l}^t, \widehat{\theta}_k \rangle) \geq$ previously best $\widetilde{U}^{T-t}(\langle \widetilde{b}_{k,l}^t, \widehat{\theta}_k \rangle)$ **then**
25.                  **if** $(\widetilde{U}_{a_k}^{T-t}(\langle \widetilde{b}_{k,l}^t, \widehat{\theta}_k \rangle) > \widetilde{U}^{T-t}(\langle \widetilde{b}_{k,l}^t, \widehat{\theta}_k \rangle)$ **then**
26.                      $\widetilde{U}^{T-t}(\langle \widetilde{b}_{k,l}^t, \widehat{\theta}_k \rangle) \leftarrow \widetilde{U}_{a_k}^{T-t}(\langle \widetilde{b}_{k,l}^t, \widehat{\theta}_k \rangle)$
27.                      OPT$(\langle \widetilde{b}_{k,l}^t, \widehat{\theta}_k \rangle) \leftarrow \phi$
28.                  OPT$(\langle \widetilde{b}_{k,l}^t, \widehat{\theta}_k \rangle) \overset{\cup}{\leftarrow} a_k$
29. **for all** $a_k \in A_k$ **do**
30.      **if** $(a_k \in$ OPT$(\langle \widetilde{b}_{k,l}^0, \widehat{\theta}_k \rangle)$ **then**
31.          $Pr(a_k | \theta_k) \leftarrow \frac{1}{|\text{OPT}(\langle \widetilde{b}_{k,l}^0, \widehat{\theta}_k \rangle)|}$
32.      **else**
33.          $Pr(a_k | \theta_k) \leftarrow 0$
34. **return** $Pr(A_k | \theta_k)$

---

Figure 18: Computing an approximately optimal finite horizon policy tree given a model containing an initial sampled belief. When $l = 0$, the exact POMDP policy tree is computed.